\documentclass[journal]{IEEEtran}
\usepackage{amsmath}   
\usepackage{lineno,hyperref} 
\usepackage{algorithmic}
\usepackage{algorithm}
\usepackage{bbding}
\usepackage{pifont}
\usepackage{wasysym}
\usepackage{amssymb}
\usepackage{color}
\usepackage{amssymb}   
\usepackage{booktabs}   
\usepackage{multicol}  
\usepackage{multirow}  
\usepackage{caption}
\usepackage{graphicx}
\usepackage{diagbox}    
\usepackage{multirow}   
\usepackage{bm} 
\usepackage{subfigure} 
\usepackage{stfloats}

\newcounter{MYtempeqncnt}

\usepackage{graphicx}
\usepackage{caption}
\usepackage{subfigure} 
\usepackage{float}  
\usepackage[subfigure]{graphfig} 
\modulolinenumbers[5]

\begin{document}


\title{Manifold-based Incomplete Multi-view Clustering via Bi-Consistency Guidance}  
\author{Huibing Wang, Mingze Yao*, Yawei Chen, Yunqiu Xu, Haipeng Liu, Wei Jia, Xianping Fu, Yang Wang,~\IEEEmembership{Senior Member, ~IEEE}
	\thanks{H. Wang, M. Yao, Y. Chen and X. Fu are with the College of Information Science and Technology, Dalian Maritime University, Liaoning, 116026, China, e-mail: (huibing.wang@dlmu.edu.cn; ymz0284@dlmu.edu.cn; cyw@dlmu.edu.cn; fxp@dlmu.edu.cn). }
	\thanks{Yunqiu Xu is with the University of Technology, Sydney, Australia, e-mail: (yunqiuxu1991@gmail.com). }
	\thanks{Haipeng Liu, Wei Jia and Yang Wang are with Hefei University of Technology, Hefei, Anhui, China, e-mail: (hpliu\_hfut@hotmail.com; jiawei@hfut.edu.cn; yangwang@hfut.edu.cn). }
	\thanks{Mingze Yao is the corresponding author.}}

\maketitle

\begin{abstract}   
	Incomplete multi-view clustering primarily focuses on dividing unlabeled data into corresponding categories with missing instances, and has received intensive attention due to its superiority in real applications. Considering the influence of incomplete data, the existing methods mostly attempt to recover data by adding extra terms. However, for the unsupervised methods, a simple recovery strategy will cause errors and outlying value accumulations, which will affect the performance of the methods. Broadly, the previous methods have not taken the effectiveness of recovered instances into consideration, or cannot flexibly balance the discrepancies between recovered data and original data. To address these problems, we propose a novel method termed \underline{M}anifold-based \underline{I}ncomplete \underline{M}ulti-view clustering via \underline{B}i-consistency guidance (MIMB), which flexibly recovers incomplete data among various views, and attempts to achieve biconsistency guidance via reverse regularization. In particular, MIMB adds reconstruction terms to representation learning by recovering missing instances, which dynamically examines the latent consensus representation. Moreover, to preserve the consistency information among multiple views, MIMB implements a biconsistency guidance strategy with reverse regularization of the consensus representation and proposes a manifold embedding measure for exploring the hidden structure of the recovered data. Notably, MIMB aims to balance the importance of different views, and introduces an adaptive weight term for each view. Finally, an optimization algorithm with an alternating iteration optimization strategy is designed for final clustering. Extensive experimental results on 6 benchmark datasets are provided to confirm that MIMB can significantly obtain superior results as compared with several state-of-the-art baselines.
	
\end{abstract}


\IEEEpeerreviewmaketitle

\section{Introduction}   
Currently, real-world data are usually extracted in assortment feature representations by multiple feature descriptors due to the rapid development of the feature extraction technology. Moreover, these assortment feature representations are termed multi-view data, and have become increasingly common \cite{peng2019comic}, \cite{wang2020discriminative}, \cite{9842337}. For example, the same piece of text paragraph in a social network can be presented either in English or in Chinese. A news item can be described as multiple representations, such as a text description, image \cite{wang2022progressive} \cite{wang2020kernelized} or video \cite{liu2024structure}. The different representations of multiple views have their own features and properties \cite{10050836, hao2022tensor}, which usually boosts the model's performance with richer information than a single view \cite{yang2017discrete}, \cite{jiang2022tensorial}. Considering the diversity of multiple views, how to effectively utilize such data and complete corresponding computer vision tasks \cite{qian2022switchable, peng2023adaptive, wang2024unpacking} is an essential but challenging endeavor.

In the field of data analysis, an increasing number of approaches and applications have been recently proposed, and multi-view clustering (MVC) \cite{wang2021survey,wang2022towards,wang2016iterative} has become a mainstream task attracting intensive attention. As a typical unsupervised machine learning algorithm, MVC tends to mostly adopt richer information among the multiple views to explore the latent clustering structures\cite{chen2021generalized,wu2019essential}, which are widely utilized in many domains. Cai et al. \cite{cai2013multi} developed a unified clustering method called the improved K-means algorithm that has multiple kernels for processing large-scale data with multiple views; this method does not require graph construction \cite{wang2023graph}, and adopts the structured sparsity-inducing norm, $l_{2,1}$-norm. Chen et al. \cite{chen2019jointly} introduced a novel nonlinear method with kernel-induced mapping and automatically learned reasonable weights for balancing the distribution of multiple views. In addition, owing to the excellent performance of the spectral clustering methods, many MVC methods attempt to integrate spectral clustering and multi-view data to explore the similarity information rather than a kernel structure for completing the clustering tasks. Kumar et al. \cite{kumar2011co} introduced two spectral clustering frameworks by coregularizing the multi-view strategy, which can mine the latent clustering information from multiple views via centroid-based and pairwise strategies. Li et al. \cite{li2021consensus} constructed a unified low-rank tensor and spectral learning framework that can learn the underlying similarity structures with a spectral embedding model and obtain the high-order correlation with a stacked tensor from multiple views. Note, that while some of these previous studies have obtained satisfactory results, the existing MVC studies mainly hypothesize that multi-view data exist without missing data. Nevertheless, in most real-life scenarios, the above hypothesis cannot be maintained due to the possibility that some instances in different views may not exist, or are lost during the collection and storage process, such as some webpages that contain only images but no text descriptions. Therefore, existing MVC methods perform poorly in clustering incomplete multi-view data, which presents a challenging task called incomplete multi-view clustering (IMVC) \cite{yin2015incomplete}.

To cope with the challenges posed by incomplete situations, many approaches have been developed for the IMVC task \cite{wen2022survey}. As the seminal contribution in the domain of incomplete multi-view clustering (IMVC), Trivedi et al. \cite{trivedi2010multiview} first introduced a method to process incomplete data, which adopted kernel canonical correlation analysis for effectively clustering incomplete data. Liu et al. \cite{liu2019multiple} proposed a simple yet effective algorithm to combine the learning of kernels and the representation for clustering into a unified framework, which designed multiple kernels to address incomplete view tasks. However, the method based on KCCA requires a complete view, i.e., at least one intact view must include all instances. Furthermore, one obvious drawback of the kernel-based methods is that they require additional prior knowledge for designing the kernel function, which can greatly influence the clustering results. Another increasingly popular approach for addressing the challenge of clustering incomplete multi-view data is factorizing the incomplete data matrix, which aims to collaboratively generate a consensus representation from multiple views. For instance, Li et al. \cite{li2014partial} explored a novel method called partial multi-view clustering (PMVC). They focused on leveraging the matrix factorization technique with a nonnegative constraint for exploring a consensus representation with partial existing instances. Under the guidance of the aforementioned studies, Xu et al. \cite{xu2018partial} introduced an integrated framework that integrates subspace representation learning and underlying structure embedding. They aimed to learn a more comprehensive data description and reduce the error problems from incomplete data. Hu et al. \cite{hu2018doubly} introduced a weighted algorithm with a seminonnegative matrix factorization method, which incorporates prior alignment information to address incomplete multi-view clustering. In addition, many graph-based methods adopt the inherent similarity structure in each view to address missing-view problems. Zhou et al. \cite{zhou2019consensus} explored a novel incomplete multi-view clustering algorithm with graph learning; they intended to obtain multiple spectral embedding matrices from multiple views and then reconstruct them as high-order tensors for constraining incomplete data. Wang et al. \cite{wang2019spectral} explored the correlation between incomplete multi-view clustering and the perturbation of spectral clustering, and verified the robustness of the spectral clustering method for incomplete mathematical data. However, these methods still have limitations in that they ignore the influence of missing views, and they assume that the existing views can preserve enough features to explore cluster structures among multiple views, which significantly impacts the clustering performance.

To suit more general incomplete situations, recovery-based incomplete multi-view clustering approaches have achieved impressive progress, which forced researchers to learn the interpolation of missing views for completing multi-view data. In early work, some approaches attempted to utilize zeros or the average values calculated by the existing instances, for recovering missing instances \cite{zhao2016incomplete,qian2023rethinking,shao2016online}. However, these native filling approaches cannot flexibly process various incomplete multi-view data, which limits their clustering performance. Therefore, recent methods have proposed more flexible algorithms for recovering missing views. For example, Wen et al. \cite{wen2019unified} introduced an extra matrix with an indicator matrix to recover incomplete data via matrix factorization in a unified framework, which integrates the reconstruction of incomplete instances and matrix factorization of the consensus representation. Yin et al. \cite{yin2021incomplete} proposed a different strategy that takes the original data into an  optimization for recovering the incomplete data of each view, and implements Laplacian regularization with NMF to explore the latent structure. Moreover, deep learning approaches have attracted intensive attention and have developed rapidly in the IMVC community \cite{xu2023adaptive}, \cite{shang2017vigan}. Specifically, Wang et al. \cite{wang2018partial} designed a robust deep learning-based framework with a cycle generative adversarial network (Cycle-GAN) to generate incomplete instances, which can also capture an effective hidden structure for complete clustering tasks simultaneously. Yang et al. \cite{yang2022robust} proposed a contrastive learning paradigm to simultaneously handle view-unaligned problems and sample-missing problems, which could be considered one of the early studies investigating the impact of noisy correspondence issues. Despite the commendable performance achieved by these previous methods, the majority of them overlooked the underlying consistency information among the multiple views, which can potentially introduce noise or even incorrect information during the process of imputing the missing data. In addition, the existing methods fail to explore the latent structure from the recovery data, and cannot effectively balance the dissimilarity distribution between the recovery data and the existing data, which impacts the robustness of the proposed model.

To overcome the aforementioned challenges, we propose MIMB, manifold-based incomplete multi-view clustering via biconsistency guidance, which is designed with biconsistency guidance for exploring the consistency information from recovered data. MIMB proposes a manifold embedding strategy for the consensus representation to explore the hidden structure of the recovered data. Specifically, MIMB factorizes the existing data matrix for learning consensus representations and explores view-specific consistency information by introducing a recovery matrix and adopting Laplacian regularization for constraints to recover missing instances. Then, MIMB proposes reverse projection regularization for the recovered data to reduce noisy, or even incorrect, information and further mine the latent consistency information. In addition, to explore the structural information between the recovered data and original data, MIMB introduces manifold embedding for the common representation from multiple views to obtain better clustering results. Simultaneously, we propose an alternating iterative optimization scheme with an adaptive weighting strategy \cite{qian2023adaptive} to cope with each variable of the objective function. Fig. \ref{whole} shows the entire MIMB procedure. In brief, MIMB makes the following important research contributions:

\begin{figure*}[tbp!]
	\centering
	\includegraphics[width=\textwidth]{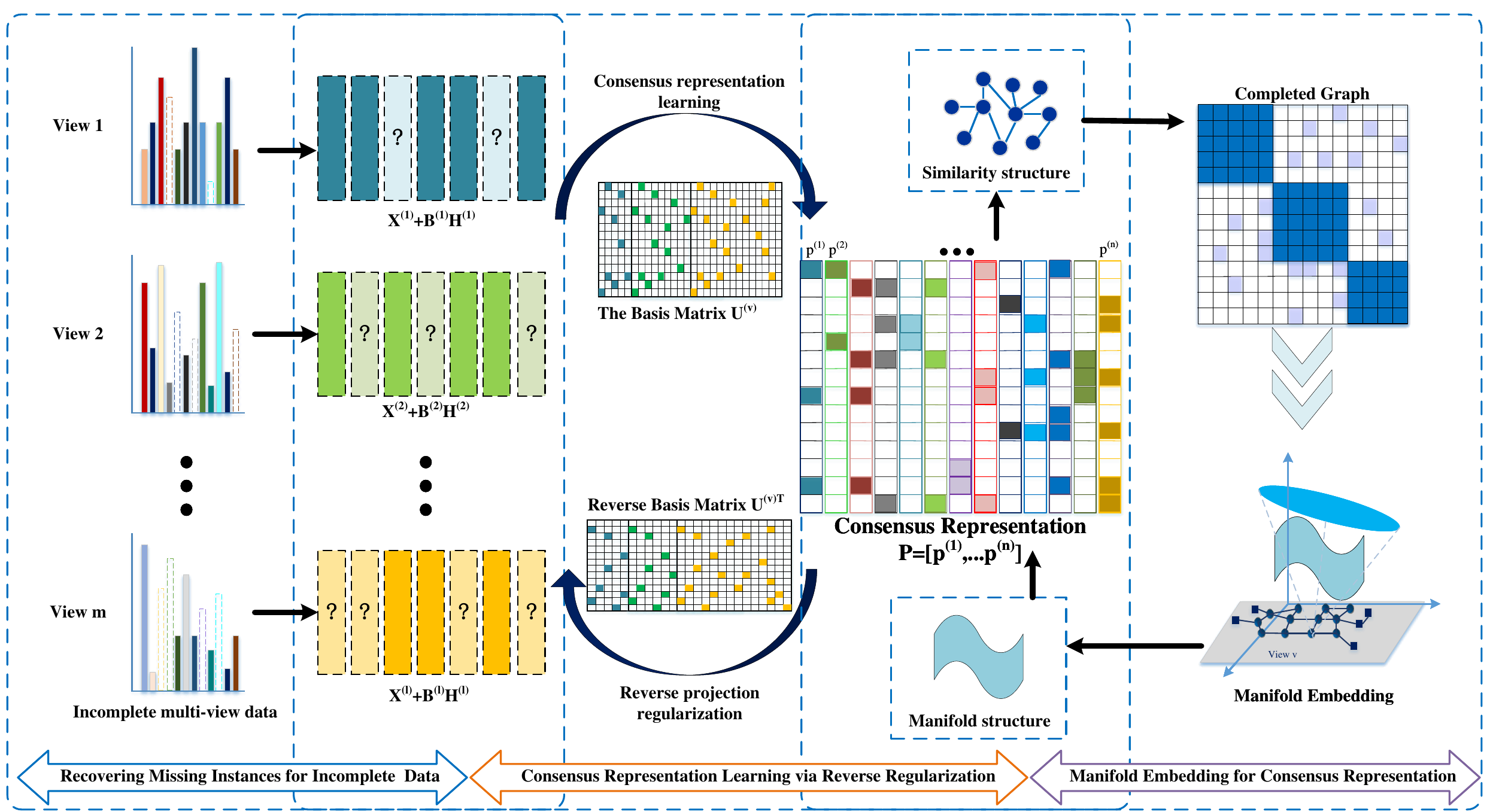} 
	\caption{The entire of manifold-based incomplete multi-view clustering via biconsistency guidance (MIMB) procedure aims to learn the consensus representation via biconsistency guidance. MIMB first recovers the missing instances for all the views. Then, MIMB utilizes biconsistency guidance for the consensus representation, which can explore the latent consistency from the recovered data. Finally, MIMB adopts manifold embedding to explore the local structure from the consensus representation for completing clustering tasks. }
	\label{whole}
\end{figure*}

\begin{itemize}
\item In this paper, a novel MIMB is proposed for incomplete multi-view clustering. The MIMB constructs a biconsistency guidance consensus representation learning framework with a recovery strategy to explore consistency information during the input process, and captures manifold structures among multiple views.
\item Different from the existing methods, MIMB takes the disparity in distributions between the recovered missing data and existing data into consideration to ensure the effectiveness of the recovered data, and explores the consistency information from multiple views with reverse regularization.
\item  A novel unified framework with manifold embedding and reverse projection regularization is developed to guarantee local structure exploration and to balance the distribution between the recovered and original data for further mining of the consistency information, which are actions that have mostly been neglected in previous works.
\end{itemize}

The remainder of this paper is organized as follows. In the second section, this paper explores some related works on incomplete multi-view clustering (IMVC) and a typical approach for IMVC. Section 3 first describes the proposed MIMB model, then presents an optimization procedure, and finally analyzes the complexity of the MIMB model. Section 4 presents extensive experimental results and analyzes the convergence of the proposed optimization algorithm to demonstrate the effectiveness of our introduced MIMB. Finally, Section 5 briefly concludes our work.

\section{Related Work}
 In this section, an overview of the existing approaches for incomplete multi-view clustering is provided first, and then, an introduction of the relevant studies related to our methods is presented. 
 
 
 \subsection{Incomplete Multi-view Clustering}
 Multi-view data always face partial-view missing data problems in real-world data collection and transmission scenarios and applications. This phenomenon leads to a novel problem, incomplete multi-view clustering (IMVC). Compared to the existing complete MVC methods, which assume that all views of each sample exist without any missing values, IMVC attempts to separate incomplete instances into matched categories with unbalanced and insufficient feature information, which is extremely difficult and challenging. In this situation, several IMVC approaches have been proposed for overcoming the partial view missing data problem. Li et al. \cite{li2014partial} introduced a novel framework called partial multi-view clustering to separate the data points of incomplete views and learn a consensus representation via low-dimensional constraints with complete data points. Xu et al. \cite{xu2018partial} introduced a clustering approach with subspace learning from multiple views for processing missing multiple views, which can learn the latent structure hidden in the underlying data space through subspace representations. However, the aforementioned methods mostly assume that a portion of the instances possess complete view information, and conduct clustering based on these instances. To solve more general incomplete situations, Liu et al. \cite{liu2021self} filled missing views as zero and adopted a self-representation subspace method for incomplete data that attempts to recover missing instances to complete multi-view clustering tasks. Wang et al. \cite{wang2019spectral} utilized average similarity values of unmissing views to complete missing similarity entries, and obtained a Laplacian matrix for spectral clustering. 
 While these imputation strategies have demonstrated the ability to handle complex structures in incomplete multi-view datasets, simple strategies may introduce additional noise and limit the clustering performance. Recently, a novel strategy for effectively completing clustering tasks with incomplete data has been introduced that attempts to recover incomplete views with existing information. Wen et al. \cite{wen2019unified} presented a combined framework for inferring missing views with embedded alignment information, which can effectively recover the missing views with an embedded matrix. With the above work, Wen et al. \cite{wen2021unified} designed a similar tensor recovery strategy, which combined manifold space-based similarity graph learning and recovered missing views. Although the above method can adaptively recover the missing views for the clustering task, the extra inferring matrix based on the original data space without any constraints will hinder the clustering task. Accurately recovering missing instances and exploring the hidden similarity structure among multiple views needs to be considered. Incorrect embeddings of feature representations can negatively impact the performance of these IMVC methods.
 
\subsection{Graph Regularized Incomplete Multi-view Clustering	}
GRIMVC exhibits similarities to methods that adopt the matrix factorization strategy with nonnegative constraints (NMFs) \cite{deng2023projective}. These methods aim to simultaneously explore the latent representation matrix while recovering incomplete multi-view data and preserving the geometrical/low-dimensional manifold structure from different views. In addition, similar to the MF-based methods, GRIMVC learns a common representation from multiple latent representation matrices. GRIMVC aims to set weights for each view and integrate them into a common representation rather than simply accumulating representation matrices. Therefore, the summarized function of GRIMVC can be provided as follows:
 \begin{gather}
\mathop {\min }\limits_{\{ {\bm{{\rm{Q}}}^{(v)}},{\bm{{\rm{R}}}^{(v)}}\} _{v = 1}^l,\bm{{\rm{P}}}} \sum\limits_{v = 1}^l 
 	\left\| {{\bm{{\rm{X}}}^{(v)}}\bm{{\rm{H}}}^{(v)} - {\bm{{\rm{Q}}}^{(v)}}{\bm{{\rm{R}}}^{(v)T}}\bm{{\rm{H}}}^{(v)}} \right\|_F^2 \\+ {\mu _v}\left\| {{\bm{{\rm{R}}}^{(v)}} - \bm{{\rm{P}}}} \right\|_F^2
 	+ \lambda Tr({\bm{{\rm{R}}}^{(v)T}}{\bm{{\rm{L}}}^{(v)}}{\bm{{\rm{R}}}^{(v)}}) \notag
 	 \\ 
s.t.\quad{\bm{{\rm{Q}}}^{(v)}} \ge 0,\hspace{2mm}{\bm{{\rm{R}}}^{(v)}} \ge 0,\hspace{2mm} \bm{{\rm{P}}} \ge 0 \notag
 \end{gather}
where the provided data in the $v$th view can be represented as $\bm{{\rm{X}}}^{(v)}\in \Re^{{d_v}\times n}$. The columns of $\bm{{\rm{X}}}^{(v)}$ are the sizes of the instances, and the rows represent the extracted feature dimensions. Moreover, the basis matrix can be denoted as $\bm{{\rm{Q}}}^{(v)}\in \Re^{d_v \times k}$, where $\bm{{\rm{R}}}^{(v)}\in \Re^{n\times k}$ represents the latent representation and $k$ represents the number of data classes. Moreover, $\bm{{\rm{H}}}^{(v)} \in \{0,1\}^{n\times n_v}$ represents the indicator matrix and is adopted for representing the incomplete instances of each view. The number of existing instances is $n_v$ in the $v$th view. The existing data of the $v$th view are reformulated as $\bm{{\rm{X}}}^{(v)}\bm{{\rm{H}}}^{(v)} \in \Re^{d_v \times n_v}$ with the above indicator matrix. $\bm{{\rm{P}}} \in \Re^{n\times k}$ denotes the consensus representation matrix, which combines every sample shared by all the views into a new representation. $\bm{{\rm{L}}}^{(v)} \in \Re^{n \times n}$ represents the Laplacian matrix and is calculated with $\bm{{\rm{W}}}^{(v)}$, which represents the similarity graph with the input data of the $v$th view. Specifically, in the normal spectral clustering method \cite{ng2001spectral}, the Laplacian matrix is usually calculated as $\bm{{\rm{L}}}^{(v)}=\bm{{\rm{I}}}-\bm{{\rm{D}}}^{(v)-1/2}\bm{{\rm{W}}}^{(v)}\bm{{\rm{D}}}^{(v)-1/2}$. $\bm{{\rm{D}}}^{(v)}$ denotes the degree matrix in which each degree is usually calculated by $\bm{{\rm{D}}}_{i,i}^{(v)}=\sum_{j = 1}{(\bm{{\rm{W}}}_{i,j}+\bm{{\rm{W}}}_{j,i})/2}$. In addition, ${\mu _v}$ and $\lambda$ denote the penalty parameters for balancing multiple views and spectral embedding, respectively. 

\section{Manifold-based Incomplete Multi-view Clustering via Bi-Consistency Guidance}
In this section, this paper is devoted to exploring effective manifold-based incomplete multi-view clustering via biconsistency guidance (MIMB), which aims to integrate the recovery of incomplete instances and the learning of the consensus representation to fully consider the consistency information among multiple views. The core of MIMB is to learn a consensus representation and obtain consistency information from the recovered data. Specifically, the entire MIMB process is displayed in Fig. \ref{whole}. To address the incomplete clustering task, MIMB first utilizes a recovery-based matrix factorization approach to simultaneously recover the incomplete instances and learn a consensus representation. Then, MIMB takes the correction between the recovered data and original data into consideration through reverse representation regularization of all the views, which can fully explore the hidden consistency information from the recovered data. In addition, MIMB integrates manifold embedding into the entire framework, which can effectively mine the local structure from the consensus representation. Table \ref{Table1} lists some descriptions of the important symbol notations. 
	 \begin{table}[tbp!]
		\centering
		\Large
		\caption{The Descriptions of Important Formula Symbols} 
		\label{Table1}
		\scalebox{0.6}{
			\begin{tabular}{c|l}
				\hline
				\multirow{2}{*}{\textbf{Notations}} & \multirow{2}{*}{\textbf{Descriptions}}                      \\
				&                                                    \\ \hline
				$\bm{{\rm{X}}}^{(v)}$      & Data matrix                     \\
				$\bm{{\rm{B}}}^{(v)}$        & Recovery matrix for recovering      \\
				$\bm{{\rm{H}}}^{(v)}$         & Index matrix for aligning incomplete data                    \\
				$\bm{{\rm{U}}}^{(v)}$        & The basis matrix from matrix factorization of $v$th view               \\
				$\bm{{\rm{P}}}$     & Consensus representation              \\
				$\bm{{\rm{S}}}$          & The manifold structure for consensus representation            \\
				$\bm{{\rm{I}}}$            & The identity matrix                                \\
				$n$                      &  Existing data size                                 \\
				$n_v^m$                    & Missing instances size\\
				$l$                        & The number of views                                 \\
				$m^v$                      & The feature dimensionality               \\
				$\alpha^v$                 & The weight parameter          \\
				\hline
		\end{tabular}}
	\end{table}

First, MIMB proposes a unique recovery-based incomplete multi-view clustering framework that utilizes consensus representation learning and inverse regularization as biconsistency guidance to recover incomplete views and to explore more consistent information. In addition, MIMB also embeds the manifold structure into the consensus representation, as shown in Eq.\ref{1}:

\begin{gather}
\label{1}
\begin{aligned}
&\mathop {\min }\limits_{\bm{{\rm{P}}},{\bm{{\rm{P}}_{1}}},{\bm{{\rm{P}}_{2}}}} \underbrace {{\cal J}\left( {\left\{ \bm{{\rm{X}}^v} \right\}_{v = 1}^l,\left\{ \bm{{\rm{P}}} \right\}} \right)}_{{\rm{Recovery-based\hspace{1mm}Representation\hspace{1mm}Learning}}}\\&+\underbrace {{{\cal L}_{\rm{1}}}\left( {\left\{ \bm{{\rm{P}_1}} \right\}} \right)}_{{\rm{Inverse\hspace{1mm}Representation\hspace{1mm}Regularization}}} + \underbrace {{{\cal L}_{\rm{2}}}\left( {\left\{ \bm{{{\rm{P}}_2}} \right\}} \right)}_{{\rm{Manifold\hspace{1mm}Embedding}}}
\end{aligned}
\end{gather}
where ${{\cal J}\left( {\left\{ \bm{{\rm{X}}^v}  \right\}_{v = 1}^l,\left\{ \bm{{\rm{P}}} \right\}} \right)}$ aims to exploit the regularized consensus representation of multiple views learned through the recovery matrix and indicator matrix of the existing views and the missing views. In addition, ${{{\cal L}_{\rm{1}}}\left( {\left\{ \bm{{\rm{P}_1}} \right\}} \right)}$ and ${{{\cal L}_{\rm{2}}}\left( {\left\{ \bm{{{\rm{P}}_2}} \right\}} \right)}$ denote inverse representation regularization and manifold embedding, respectively, which are introduced on the consensus representation $ \bm{{\rm{P}}} $ to further explore more consistency information through all recovered incomplete views by imposing reverse regularization of $ \bm{{\rm{P}}} $ and manifold embedding. In addition, based on the recovery item, MIMB can mine an ideal latent consensus representation $ \bm{{\rm{P}}} $ from multiple recovered views, which can effectively mine consistent and complementary information. Moreover, MIMB encourages the use of reverse regularization constraints and manifold embedding on the learned representation $ \bm{{\rm{P}}} $, which explores the local structure and balances the discrepancy distribution between the recovered data and the original data. To fully utilize different views, the proposed framework (as shown in Eq. \ref{1}) also introduces an adaptive weight term for each view to balance different views and encourage the mining of manifold structures from recovered data.
\subsection{Consensus Representation Learning with Recovery Regularization}
Most normal NMF-based multi-view clustering approaches learn an effective consensus representation among diverse views. However, the existing works almost learn the consensus representation by utilizing the available instances, which ignores the influence of incomplete data. Therefore, we designed the following model, which fully utilizes the existing instances and introduces the recovery matrix for consensus representation learning:
\begin{gather}
\label{3}	
\mathop {\min }\limits_{\bm{{\rm{B}}}^{(v)},\bm{{\rm{U}}}^{(v)},\bm{{\rm{P}}}} \sum\limits_{v = 1}^l {\left( \begin{array}{l} {\left\| {{\bm{{\rm{X}}}^{(v)}} + {\bm{{\rm{B}}}^{(v)}}{\bm{{\rm{H}}}^{(v)}} - \bm{{\rm{U}}}^{(v)}\bm{{\rm{P}}}} \right\|_F^2 }\\{+ \frac{{{\lambda _1}}}{2}\sum\limits_{j = 1}^{m_{v}} {\sum\limits_{i = 1}^{m_{v}} {\left\| {\bm{{\rm{B}}}_{i,:}^{(v)} - \bm{{\rm{B}}}_{j,:}^{(v)}} \right\|_2^2 \bm{{\rm{G}}}_{i,j}^{(v)}} } } \end{array} \right)}  \\
s.t.\quad {\bm{{\rm{U}}}^{(v)T}}{\bm{{\rm{U}}}^{(v)}} = \bm{{\rm{I}}} \notag
\end{gather}
where ${\bm{{\rm{B}}}^{(v)}} \in {\Re^{{m_v} \times n_v^m}}$ and ${\bm{{\rm{H}}}^{(v)}} \in {\Re^{n_v^m \times n }}$ denote the recovery matrix and index matrix, respectively, which are utilized to recover the missing instances for the proposed MIMB. Specifically, zeros are filled in the incomplete data for representing the missing samples, which have a size of ${n_v^m}$ in the $v$th view. In addition, MIMB adopts graph structure embedding on the recovery matrix to mine the essential information, where ${\bm{{\rm{B}}}_{i,:}^{( v)}}$ and ${\bm{{\rm{B}}}_{j,:}^{(v)}}$ denote the $i$th and $j$th row vectors of ${\bm{{\rm{B}}}^{(v)}}$, respectively, and $\lambda _1 $ denotes the positive penalty parameter. Moreover, with the matrix factorization model, the basis matrix is represented as ${\bm{{\rm{U}}}^{(v)}} \in {\Re^{{m_v} \times c}}$, and the consensus representation is denoted as $\bm{{\rm{P}}} \in {\Re^{c \times n}} $, where $c$ represents the class label numbers. In general, the index matrix ${\bm{{\rm{H}}}^{(v)}}$ of the $v$th view needs to be predefined as:
\begin{gather}
\label{index}
\bm{{\rm{H}}}_{i,j}^{v} = \left\{ {\begin{array}{l}
	1,\quad if\hspace{1mm} the\hspace{1mm} $j$th\hspace{1mm} instance\hspace{1mm} is\hspace{1mm} missing\hspace{1mm} instance\\
	0,\quad otherwise
	\end{array}} \right.
\end{gather}

In addition, in Eq. \ref{3}, the neighbor graph $\bm{{{\rm{G}}}^{(v)}}\in {\Re^{{m_v} \times {m_v}}}$ is preconstructed as described above. It is easy to translate Problem Eq. \ref{3} into the following formula:

\begin{gather}
\label{5}	
\mathop {\min }\limits_{\bm{{{\rm{B}}}^{(v}},\bm{{{\rm{U}}}^{(v)}},\bm{{\rm{P}}}} \sum\limits_{v = 1}^l {\left( \begin{array}{l}{\left\| {\bm{{\rm{X}}}^{(v)} + {\bm{{\rm{B}}}^{(v)}}{\bm{{\rm{H}}}^{(v)}} - {\bm{{\rm{U}}}^{(v)}}\bm{{\rm{P}}}} \right\|_F^2}\\{  + {\lambda _1}Tr\left( {{\bm{{\rm{B}}}^{(v)T}}{\bm{{\rm{L}}}^{(v)}}{\bm{{\rm{B}}}^{(v)}}} \right)}   \end{array} \right)}\\
s.t.\quad {\bm{{\rm{U}}}^{(v)T}}{\bm{{\rm{U}}}^{(v)}}= \bm{{\rm{I}}} \notag
\end{gather}
where ${{\bm{{\rm{L}}}^{(v)}}}$ denotes the Laplacian matrix, which is calculated by the preconstructed matrix ${\bm{{\rm{G}}}^{(v)}}$ as ${\bm{{\rm{L}}}^{(v)}} = {\bm{{\rm{D}}}^{(v)}} -{\bm{{\rm{G}}}^{(v)}}$. $\bm{{\rm{D}}}_{i,j}^{(v)} = \sum\nolimits_{j = 1}^{{m_v}} {\bm{{\rm{G}}}_{i,j}^{(v)}}$ is usually calculated for the diagonal matrix ${\bm{{\rm{D}}}^{( v )}}$. In Eq. \ref{5}, the MIMB adopts the orthogonal constraint ${\bm{{\rm{U}}}^{(v)T}}{\bm{{\rm{U}}}^{^{(v)}}}=\bm{{\rm{I}}}$ to ensure that $\bm{{\rm{U}}}$ is independent during optimization.

According to Eq. \ref{5}, ${{\bm{{\rm{X}}}^{( v )}}+{\bm{{\rm{E}}}^{( v )}}}{\bm{{\rm{W}}}^{( v )}}$ is the recovered multi-view data. By recovering incomplete multi-view data, MIMB can flexibly learn the consensus representation among the complete views. Moreover, the graph regularization term, which aims to effectively guide the proposed MIMB to learn the consensus representation from the latent space, has been introduced to reduce errors in the recovery matrix.

\subsection{Reverse Consensus Representation}
It is crucial to explore the consistency information among multiple views and mine the relationships between the recovered instances and existing instances, which is the main bottleneck for incomplete multi-view clustering tasks. Therefore, our proposed MIMB imposes a reverse consensus representation regularization constraint on the recovered data to fully explore the consistency information and balance the discrepancy distribution from all the views. Specifically, we utilize reverse projection for consensus representation learning and regularize the basis matrix, and the framework is as follows:
\begin{gather}
\label{6}
\mathop {\min }\limits_{{\bm{{\rm{U}}}^{(v)}},\bm{{\rm{P}}}} \sum\limits_{v = 1}^l {\left( {\left\| {{\bm{{\rm{U}}}^{(v)T}}\left( {{\bm{{\rm{X}}}^{(v)}} + {\bm{{\rm{B}}}^{(v)}}{\bm{{\rm{H}}}^{(v)}}} \right)-\bm{{\rm{P}}}} \right\|_F^2 + \beta \left\| {{\bm{{\rm{U}}}^{(v)}}} \right\|_F^2} \right)} \\
s.t.\quad {\bm{{\rm{U}}}^{(v)T}}{\bm{{\rm{U}}}^{(v)}} =  \bm{{\rm{I}}} \notag
\end{gather}
where the reverse regularization term is denoted as ${{\bm{{\rm{U}}}^{(v)T}}\left( {{\bm{{\rm{X}}}^{(v)}} + {\bm{{\rm{B}}}^{(v)}}{\bm{{\rm{H}}}^{(v)}}} \right)}$, which aims to reverse project the recovered multi-view data to the consensus representation for exploring the consistency information among multiple views. Moreover, the above term can also fully consider the relationship between the existing data and the recovered data. Moreover, the Frobenius norm of the basis matrix ${{\bm{{\rm{U}}}^{(v)}}}$ is proposed to avoid trivial solutions, and the parameter $\beta $ is introduced to balance the constraints of the regular term in the overall objective function.

\begin{figure*}[bp!]
	\normalsize
	\setcounter{MYtempeqncnt}{\value{equation}}
	\setcounter{equation}{9}
	\begin{gather}
	\label{10}
	\begin{aligned}
	&\mathop {\min }\limits_{\bm{{\rm{B}}}^{(v)},\bm{{\rm{U}}}^{(v)},\bm{{\rm{P}}},\bm{{\rm{S}}},\bm{{\rm{F}}},\alpha^{(v)}}  \sum\limits_{v = 1}^l \underbrace {{{( {\alpha ^{(v)}} )}^r} ( \| {{\bm{{\rm{Y}}}^{(v)}} - {\bm{{\rm{U}}}^{(v)}}\bm{{\rm{P}}}} \|_F^2 + {\lambda _1}Tr( {{\bm{{\rm{B}}}^{(v)T}}{\bm{{\rm{L}}}^{(v)}}{\bm{{\rm{B}}}^{(v)}}} )}_{{\rm{Recovery-based\hspace{1mm}Representation\hspace{1mm}Learning}}} + \underbrace{\| {{\bm{{\rm{U}}}^{(v)T}}{\bm{{\rm{Y}}}^{(v)}} - \bm{{\rm{P}}}} \|_F^2 + \beta \| {{\bm{{\rm{U}}}^{(v)}}} \|_F^2 )}_{{\rm{Inverse\hspace{1mm}Representation\hspace{1mm}Regularization}}} \\&+\underbrace{ {\lambda _2}\| {{\bm{{\rm{S}}} - {\bm{{\rm{P}}}^T}\bm{{\rm{P}}}} \|}_F^2 +  {\lambda _3}Tr( {{\bm{{\rm{F}}}^T}{\bm{{\rm{L}}}_S}\bm{{\rm{F}}}} )}_{{\rm{Manifold\hspace{1mm}Embedding}}}\\
	&s.t.\quad {\bm{{\rm{U}}}^{(v)T}}{\bm{{\rm{U}}}^{(v)}}  =  \bm{{\rm{I}}},\hspace{2mm}\sum\limits_{v = 1}^l {{( {{\alpha ^{(v)}}} )}^r} = { 1,}\hspace{2mm}  {\alpha ^{(v)}}\ge 0 ,\hspace{2mm} {\forall _{i,}}{\bm{{\rm{S}}}_{i,:}}1 =  1 ,0 \le {\bm{{\rm{S}}}_{i,j}} \le 1 ,\hspace{2mm} {\bm{{\rm{S}}}_{i,i}} = 0,\hspace{2mm}{\bm{{\rm{F}}}^T}\bm{{\rm{F}}} = \bm{{\rm{I}}} 
	\end{aligned}
	\end{gather}
	\setcounter{equation}{\value{MYtempeqncnt}}
	\hrulefill
	\vspace*{4pt}
\end{figure*}

\subsection{Manifold Embedding for Consensus Representation}
Most existing multi-view datasets naturally share common local structures and clustering information across multiple views due to their properties. Manifold structure preservation is indispensable, and plays a vital role in completing multi-view clustering tasks, which aim to simultaneously explore the common structure from recovered data and improve the compactness of the consensus representation. However, it is clearly impossible to explore the hidden manifold structure with incomplete data since the recovered data are unbalanced from the complete data. Therefore, to solve the above problem, MIMB proposes a manifold embedding learning model:
\begin{gather}
\label{7}
\begin{aligned}
& \mathop {\min }\limits_{\bm{{\rm{P}}},\bm{{\rm{S}}}} {\lambda _2}\left\| {\left. {\bm{{\rm{S}}} - {\bm{{\rm{P}}}^T}\bm{{\rm{P}}}} \right\|} \right._F^2 \\
& s.t.\quad \forall i,{\bm{{\rm{S}}}_{i,:}}1= 1,\quad0 \le {\bm{{\rm{S}}}_{i,j}} \le 1
\end{aligned}
\end{gather}
where the nearest neighbor graph is represented as $\bm{{\rm{S}}}\in {\Re^{n \times n}}$. Each element of the graph is calculated as the degree of similarity between the corresponding two instances. To avoid situations where no sample is connected to its adjacent samples, MIMB introduces the constraint ${\bm{{\rm{S}}}_{i,:}}1=1$ to ensure the reliability of the sample similarity degrees.

In addition, MIMB adds a rank constraint on $\bm{{\rm{S}}}$ to obtain a better clustering performance, which can be represented in Eq. \ref{7} as follows:
\begin{gather}
\begin{aligned}
\label{8}
&\mathop {\min }\limits_{\bm{{\rm{P}}},\bm{{\rm{S}}},\bm{{\rm{F}}}} {\lambda _2}\left\| {\left. {\bm{{\rm{S}}} - {\bm{{\rm{P}}}^T}\bm{{\rm{P}}}} \right\|} \right._F^2 + {\lambda _3}Tr\left( {{\bm{{\rm{F}}}^T}{\bm{{\rm{L}}}_S}\bm{{\rm{F}}}} \right)\\
&s.t.\quad \forall i,{\bm{{\rm{S}}}_{i,:}}1 = 1,\quad 0 \le {\bm{{\rm{S}}}_{i,j}} \le 1 ,\quad {\bm{{\rm{F}}}^T}\bm{{\rm{F}}} = \bm{{\rm{I}}}
\end{aligned}
\end{gather}

By embedding the manifold structure into the learned consensus representation, the consistency information from multiple completed views can be guaranteed for a more comprehensive exploration. In conclusion, the proposed MIMB recovers incomplete views and learns the consensus representation to obtain the final clustering results.

\subsection{Adaptively Weighted for MIMB}

Because multi-view data have different information among the diverse views, these various views typically exhibit different physical meanings and contain different feature information due to their diverse views. Particularly in the context of incomplete multi-view clustering, the presence of incomplete views can lead to an imbalance issue among the available instances across multiple views. The uncertain situations of incomplete data seriously reduce the clustering performances of the existing methods. Therefore, MIMB adopts an adaptive weighted term for each view to balance the importance:
\begin{gather}
\label{9}
\mathop {\min }\limits_{{\alpha ^{(v)}}} \sum\limits_{v = 1}^l {{{\left( {\alpha ^{(v)}} \right)}^r}} {\Lambda ^{(v)}}\qquad 
\\s.t.\quad\sum\limits_{v = 1}^l {{{\left( {{\alpha ^{(v)}}} \right)}^r}  =  1,} \hspace{2mm} {\alpha ^{( v )}} \ge 0 \notag
\end{gather}
where ${\Lambda ^{\left( v \right)}}$ represents the remaining function terms of the proposed biconsistency guidance model. In addition, the positive weight ${{\alpha ^{\left( v \right)}}}$ of the $v$th view is introduced to balance the significance of the different views. We utilize the constraint $r > 1$ to ensure the smoothness of the weight distribution. As a result, MIMB can dynamically address the synergistic effects of different views in the optimization process, which is critical for multi-view clustering.

\subsection{Overall Objective Function of the MIMB}
Consequently, the overall objective function of the MIMB based on all the above considerations can be organized as Eq. \ref{10}:
where ${\bm{{{\rm{Y}}}}^{(v)}} = {\bm{{\rm{X}}}^{(v)}} + {\bm{{\rm{B}}}^{(v)}}{\bm{{\rm{H}}}^{(v)}}$, and the penalty parameters ${\lambda _2}, \lambda_3$ are introduced to balance the contributions of the different parts.

\subsection{Optimization Procedure}
In this section, our proposed MIMB optimization procedure is described in detail. The objective function is iteratively optimized by decomposing the problem into multiple steps. In each step of the optimization procedure, a corresponding variable is updated, while the other variables are kept fixed until objective convergence.

$\bm{{\rm{Updating}} }\quad \bm{{\rm{P}}}$: The MIMB first considers the other variables as constants, and the solution of Eq. \ref{10} with $\bm{{\rm{P}}}$ can be rewritten as follows:

\begin{gather}
\setcounter{equation}{10} 
\label{11}
\begin{aligned}
\Gamma  = &\sum\limits_{v = 1}^l {{{( {\alpha ^v} )}^r}} ( {\| {{\bm{{\rm{Y}}}^{(v)}} - {\bm{{\rm{U}}}^{(v)}}\bm{{\rm{P}}}} \|_F^2 + \| {{\bm{{\rm{U}}}^{(v)T}}{\bm{{\rm{Y}}}^{(v)}} - \bm{{\rm{P}}}} \|_F^2} ) \\&+ {\lambda _2}\| {{\bm{{\rm{S}}} - {\bm{{\rm{P}}}^T}\bm{{\rm{P}}}} \|}_F^2
\end{aligned}
\end{gather}

To address the above problem, the derivative of $\Gamma$ with respect to $\bm{{\rm{P}}}$ can be expressed as follows:
\begin{gather}
\begin{aligned}
\frac{{\partial \Gamma }}{{\partial P}} =& ( {{\lambda _2}\bm{{\rm{P}}}{\bm{{\rm{P}}}^T}\bm{{\rm{P}}} + \sum\limits_{v = 1}^l {{{( {{\alpha ^v}} )}^r}\bm{{\rm{P}}}} } ) \\&- ( {\sum\limits_{v = 1}^l {{{( {{\alpha ^v}} )}^r}{\bm{{\rm{U}}}^{(v)T}}{\bm{{\rm{Y}}}^{(v)}} + {\lambda _2}\bm{{\rm{PS}}}} } )
\end{aligned}
\end{gather}
Because the above solution of $\bm{{\rm{P}}}$ cannot be solved by setting the derivation to zero, we provide an appropriate method for obtaining $\bm{{\rm{P}}}$ as follows:
\begin{gather}
\bm{{\rm{P}}} \leftarrow \bm{{\rm{P}}} - {\eta _p} \odot \frac{{\partial \Gamma }}{{\partial \bm{{\rm{P}}}}}
\end{gather}
where fully considering $\bm{{\rm{P}}}$ is nonnegative. Then, the appropriate learning rate ${\eta _P}$ is introduced to ensure the nonnegativity of the proposed method in each iteration. Setting ${\eta _P} = \bm{{\rm{P}}}/({\lambda _2}\bm{{\rm{P}}}{\bm{{\rm{P}}}^T}\bm{{\rm{P}}} + \sum\limits_{v = 1}^l {{{({\alpha ^v})}^r}\bm{{\rm{P}}}} )$ yields:

\begin{gather}
\begin{aligned}
\label{12}
\bm{{\rm{P}}} \leftarrow \bm{{\rm{P}}} \odot \frac{{\sum\limits_{v = 1}^l {{{( {{\alpha ^v}} )}^r}{\bm{{\rm{U}}}^{(v)T}}{\bm{{\rm{Y}}}^{(v)}} + {\lambda _2}\bm{{\rm{PS}}}} }}{{{\lambda _2}\bm{{\rm{P}}}{\bm{{\rm{P}}}^T}\bm{{\rm{P}}} + \sum\limits_{v = 1}^l {{{( {{\alpha ^v}} )}^r}\bm{{\rm{P}}}} }}
\end{aligned}
\end{gather}

$\bm{{\rm{Updating}} }\quad  \bm{{\rm{S}}}$: From Eq. \ref{10}, the subproblem for variable $\bm{{\rm{S}}}$ can be optimized by minimizing the following equation:
\begin{gather}
\begin{aligned}
\label{13}
& \mathop {\min }\limits_{\bm{{\rm{S}}}} {\lambda _2}\| {{\bm{{\rm{S}}} - {\bm{{\rm{P}}}^T}\bm{{\rm{P}}}} \|} _F^2 + {\lambda _3}Tr( {{\bm{{\rm{F}}}^T}{\bm{{\rm{L}}}_S}\bm{{\rm{F}}}} )\\ 
&s.t.\quad {\forall _{i,}}{\bm{{\rm{S}}}_{i,:}}1  = 1 ,\hspace{2mm}0 \le {\bm{{\rm{S}}}_{i,j}} \le 1, \hspace{2mm} {\bm{{\rm{S}}}_{i,i}} = 0
\end{aligned}
\end{gather}

Considering the solution for the matrix trace, the above equation can be simplified as follows:
\begin{gather}
\begin{aligned}
\label{18}
\mathop {\min }\limits_{\bm{{\rm{S}}}} {\lambda _2}\sum\limits_{i,j;i \ne j}^n {{{( {{s_{ij}} - {a_{ij}}} )}^2}}  + \frac{{{\lambda _3}}}{{\rm{4}}}\bm{{\rm{H}}}_{i,j}^F{s_{ij}}
\end{aligned}
\end{gather}
where $\bm{{\rm{H}}}_{i,j}^F = \sum\limits_{i,j}^n {\| {{{\bm{{\rm{F}}}_{i,;}} - {\bm{{\rm{F}}}_{j,;}}} \|}_2^2}$ and ${a_{ij}} = {( {{\bm{{\rm{P}}}^T}\bm{{\rm{P}}}} )_{ij}}$, which are expanded to the sample level. The closed-form solution of problem Eq. \ref{18} can be achieved by an efficient algorithm proposed in \cite{nie2016constrained}.

$\bm{{\rm{Updating}} }\quad \bm{{\rm{F}}}$: With the optimization of $\bm{{\rm{F}}}$, MIMB regards the other variables as constants. Therefore, the optimization function of Eq. \ref{10} is reformulated as:
\begin{gather}
\begin{aligned}
\label{17}
&\mathop {\min }\limits_{\bm{{\rm{F}}}} Tr({\bm{{\rm{F}}}^T}{\bm{{\rm{L}}}_S}\bm{{\rm{F}}})\\
&s.t.\quad{\bm{{\rm{F}^T}\bm{{\rm{F}}}}} = \bm{{\rm{I}}}
\end{aligned}
\end{gather}
which can be optimized using a set of eigenvectors associated with the $c$ smallest eigenvalues of ${\bm{{\rm{L}}}_S}$.

$\bm{{\rm{Updating}} }\quad\bm{{{\rm{U}}^{(v)}}}$: By considering that all the other variables are constants, the optimization problem for computing variable ${{\bm{{\rm{U}}}^{(v)}}}$ is formulated as follows:
\begin{gather}
\begin{aligned}
\label{14}
\begin{array}{l}
\mathop {\min }\limits_{{\bm{{\rm{U}}}^{(v)}}} \| {{\bm{{\rm{Y}}}^{(v)}} - {\bm{{\rm{U}}}^{(v)}}\bm{{\rm{P}}}} \|_F^2 + \| {{\bm{{\rm{U}}}^{(v)T}}{\bm{{\rm{Y}}}^{(v)}} - \bm{{\rm{P}}}} \|_F^2 + \beta \| {{\bm{{\rm{U}}}^{(v)}}} \|_F^2\\
s.t.\quad{\bm{{\rm{U}}}^{(v)T}}{\bm{{\rm{U}}}^{(v)}} = I
\end{array}
\end{aligned}
\end{gather}
where we define ${\bm{{{\rm{Y}}}}^{(v)}} = {\bm{{\rm{X}}}^{(v)}} + {\bm{{\rm{B}}}^{(v)}}{\bm{{\rm{H}}}^{(v)}}$ to represent the recovered data. Therefore, the MIMB can obtain the optimized results as ${\bm{{\rm{U}}}^{(v)}}=\bm{{\rm{T}}}{\bm{{\rm{R}}}^T}$, where $\bm{{\rm{T}}}$ and $\bm{{\rm{R}}}$ express the results from the SVD algorithm (i.e., left singular and right singular). 

$\bm{{\rm{Updating}} }\quad{\bm{{\rm{B}}}^{(v)}} $: Keeping the consideration that the other variables are constants, the optimized results of the solution in problem Eq. \ref{10} are:
\begin{gather}
\begin{aligned}
\label{15}
&\mathop {\min }\limits_{\bm{{\rm{B}}}^{(v)}} \| {\bm{{\rm{Y}}}^{(v)}} - {\bm{{\rm{U}}}^{(v)}}\bm{{\rm{P}}} \|_F^2 + {\lambda _1}Tr( {{\bm{{\rm{B}}}^{(v)T}}{\bm{{\rm{L}}}^{(v)}}{\bm{{\rm{B}}}^{(v)}}} ) \\&+ \| {{\bm{{\rm{U}}}^{(v)T}}{\bm{{\rm{Y}}}^{(v)}} - \bm{{\rm{P}}}} \|_F^2
\end{aligned}
\end{gather}

Since ${\bm{{\rm{B}}}^{(v)}}$ denotes the recovery results, the corresponding instances in the incomplete $\bm{{\rm{X}}}^{(v)}$ are zeros. Therefore, we have rewritten Eq. \ref{15} as the following formula:
\begin{gather}
\label{20-1}
\begin{aligned}
{\cal L} ({\bm{{\rm{B}}}^{(v)}}) =& \left\| {{\bm{{\rm{B}}}^{(v)}} - {\bm{{\rm{U}}}^{(v)}}\bm{{\rm{P}}}{\bm{{\rm{W}}}^{(v)T}}} \right\|_F^2 + {\lambda _1}Tr({\bm{{\rm{B}}}^{(v)T}}{\bm{{\rm{L}}}^{(v)}}{\bm{{\rm{B}}}^{(v)}}) \\&+ \left\| {{\bm{{\rm{U}}}^{(v)T}}{\bm{{\rm{B}}}^{(v)}} - \bm{{\rm{P}}}{\bm{{\rm{W}}}^{(v)T}}} \right\|_F^2
\end{aligned}
\end{gather}

Obviously, the optimal solution can be obtained by setting the partial derivation ${{\partial \phi ( {{\bm{{\rm{B}}}^{(v)}}} )} \mathord{/
		{\vphantom {{\partial \phi ( {{\bm{{\rm{B}}}^{(v)}}} )} {\partial {\bm{{\rm{B}}}^{(v)}}}}}
		\kern-\nulldelimiterspace} {\partial {\bm{{\rm{B}}}^{(v)}}}} = 0$, then Eq. \ref{20-1} is solved as:
\begin{gather}
\label{19}
{\bm{{\rm{B}}}^{(v)}} = {( {2\bm{{\rm{I}}} + {\lambda _1}{\bm{{\rm{L}}}^{(v)}}} )^{ - 1}}{\bm{{\rm{U}}}^{(v)}}\bm{{\rm{P}}}{\bm{{\rm{W}}}^{(v)T}}\
\end{gather}

$\bm{{\rm{Updating}} }\quad{\alpha ^{(v)}} $: The balance weight ${\alpha ^{(v)}}$ of each view can be obtained by minimizing the objective problem while fixing the other variables:
\begin{gather}
\label{16}
\mathop {\min }\limits_{{\alpha ^{(v)}} > 0,\sum\limits_{v = 1}^l {{\alpha ^{(v)}} = 1} } \sum\limits_{{\rm{v}} = 1}^l {{{( {{\alpha ^{(v)}}} )}^r}{\Lambda^{(v)}}} 
\end{gather}
where ${\Lambda}^{(v)} = \| \bm{{\rm{Y}}}^{(v)} - \bm{{\rm{U}}}^{(v)}\bm{{\rm{P}}} \|_F^2 + {\lambda _1}Tr( {{\bm{{\rm{B}}}^{(v)T}}{\bm{{\rm{L}}}^{(v)}}{\bm{{\rm{B}}}^{(v)}}} ) + \| {{\bm{{\rm{U}}}^{(v)T}}{\bm{{\rm{Y}}}^{(v)}} - \bm{{\rm{P}}}} \|_F^2 + \beta \| {{\bm{{\rm{U}}}^{(v)}}} \|_F^2$. The optimal solution of $\alpha ^{(v)}$ to Eq. \ref{16} is given as follows:
\begin{gather}
\label{20}
{\alpha ^{\left(v \right)}} = {\raise0.7ex\hbox{${{{\left( {{\Lambda^v}} \right)}^{{1 \mathord{\left/
							{\vphantom {1 {1 - r}}} \right.
							\kern-\nulldelimiterspace} {1 - r}}}}}$} \!\mathord{\left/
		{\vphantom {{{{\left( {{\Lambda^v}} \right)}^{{1 \mathord{\left/
									{\vphantom {1 {1 - r}}} \right.
									\kern-\nulldelimiterspace} {1 - r}}}}} {\sum\limits_{v = 1}^l {{{\left( {{\Lambda^{(v)}}} \right)}^{{1 \mathord{\left/
										{\vphantom {1 {1 - r}}} \right.
										\kern-\nulldelimiterspace} {1 - r}}}}} }}}\right.\kern-\nulldelimiterspace}
	\!\lower0.7ex\hbox{${\sum\limits_{v = 1}^l {{{\left( {{\Lambda^{(v)}}} \right)}^{{1 \mathord{\left/
								{\vphantom {1 {1 - r}}} \right.
								\kern-\nulldelimiterspace} {1 - r}}}}} }$}}
\end{gather}

The complete optimization process of the overall objective function (Eq. \ref{10}) has been presented in the above descriptions, and Algorithm \ref{alg:B} also summarizes this process, which aims to iteratively update the different variables of the proposed MIMB until convergence. In each iteration, the calculation of $\bm{{\rm{F}}}$ is the most time-consuming part, which requires ${{O}}( {{{{n}}^3}})$. In addition, the update to $\bm{{\rm{P}}}$ requires $O( {l{n^2}} )$, and $\bm{{{\rm{U}}}}^{(v)}$ requires ${{O}}({\sum\nolimits_{{{v = 1}}}^{{l}} {{{{m}}_{{v}}}{{{c}}^{{2}}}} })$. Moreover, the optimization of ${\bm{{\rm{B}}}^{(v)}}$ requires the inverse operation and requires ${( {2{\rm{I}} + {\lambda _1}{L^{(v)}}} )^{ - 1}}$, which can be ignored because it consumes the least time. In the last step, $i.e.$, the balance weight ${{\rm{\alpha }}^{\left( v \right)}}$-step is updated, and the solution of Eq. \ref{20} can be efficiently computed via a numerical division operation.
As a result, the total time complexity of Algorithm \ref{alg:B} is $O( {\tau( {l{n^2} + {n^3} + \sum\nolimits_{v = 1}^l {{m_v}{c^2}}})} )$, where $\tau $ denotes the number of iterations.

\begin{algorithm}[tbp!] 
	\caption{MIMB Algorithm}  
	\label{alg:B}  
	\begin{algorithmic}[1] 
		\REQUIRE {Incomplete multi-view data $\bm{{\rm{X}}} = \{ {{\bm{{\rm{X}}}^{(v)}}} \}_{v = 1}^l$ with filling zeros to the missing instances, the index matrix $\bm{{\rm{H}}} = \{ {{\bm{{\rm{H}}}^{(v)}}} \}_{v = 1}^l$, parameters ${\lambda _1}$, ${\lambda _2}$, ${\lambda _3}$, $r$, pre-constructed graph $\bm{{\rm{G}}} = \{ {{\bm{{\rm{G}}}^{(v)}}} \}_{v = 1}^l$. } 
		\STATE Initialization: ${\alpha ^{(v)}}= {1 /\ l}$, the basis matrix ${\bm{{\rm{U}}}^{(v)}}$ with random values, the recovery matrix with random values ${\bm{{\rm{B}}}^{(v)}}$, random graph $\bm{{\rm{S}}}$, initialize $\bm{{\rm{F}}}$ by solving Eq. \ref{17}.
		\WHILE {not converged}
		\STATE Update consensus representation $\bm{{\rm{P}}}$ by using Eq. \ref{12};
		\STATE Update the manifold graph $\bm{{\rm{S}}}$ by solving problem Eq. \ref{18};
		\STATE Update the Laplacian representation $\bm{{\rm{F}}}$ using Eq. \ref{17};
		\FOR{$v$ from $1$ to $l$}
		\STATE Update basis matrix ${{\bm{{\rm{U}}}^{(v)}}}$ by minimizing Eq. \ref{14};
		\STATE Update recovering matrix ${\bm{{\rm{B}}}^{(v)}}$ by solving Eq. \ref{19};
		\ENDFOR
		\STATE Update the balance weight ${\alpha ^{(v)}}$ according to Eq. \ref{20}
		\ENDWHILE
		\ENSURE  ${\bm{{\rm{U}}}^{(v)}}, {\bm{{\rm{B}}}^{(v)}}, \bm{{\rm{P}}}, \bm{{\rm{S}}}$;    
	\end{algorithmic}  
\end{algorithm}
\begin{table*}[tbp!]
	\centering
	\caption{Mean values of performance with various methods on the BBCSport, 3Sources and ORL datasets with different incomplete ratios.} 
	\label{Table2}
	\scalebox{1}{
		\begin{tabular}{cc|ccc|ccc|ccc}
			\hline
			\multicolumn{2}{c|}{}                                                   & \multicolumn{3}{c|}{ACC} & \multicolumn{3}{c|}{NMI} & \multicolumn{3}{c}{Purity} \\ \hline
			\multicolumn{1}{c|}{Datasets}                   & Method/\ Missing rate & 0.1    & 0.3    & 0.5    & 0.1     & 0.3    & 0.5   & 0.1    & 0.3     & 0.5     \\ \hline
			\multicolumn{1}{c|}{\multirow{10}{*}{BBCSport}} & BSV                   & 58.62  & 51.31  & 44.03  & 43.73   & 31.03  & 21.40 & 65.79  & 55.07   & 47.59   \\
			\multicolumn{1}{c|}{}                           & Concat                & 70.62  & 58.72  & 33.21  & 61.69   & 38.92  & 18.61 & 80.59  & 63.24   & 37.00   \\
			\multicolumn{1}{c|}{}                           & DAIMC                 & 68.62  & 63.45  & 56.89  & 56.62   & 50.17  & 37.89 & 76.90  & 71.72   & 61.03   \\
			\multicolumn{1}{c|}{}                           & IMSC-AGL              & 54.14  & 52.93  & 45.69  & 35.66   & 31.56  & 21.75 & 58.28  & 56.72   & 50.86   \\
			\multicolumn{1}{c|}{}                           & OPIMC                 & 76.41  & 74.48  & 69.31  & 70.46   & 66.11  & 54.57 & 87.41  & 85.00   & 78.10   \\
			\multicolumn{1}{c|}{}                           & PIC                   & 76.72  & 72.41  & 58.62  & 61.06   & 68.17  & 45.67 & 76.72  & 84.48   & 66.38   \\
			\multicolumn{1}{c|}{}                           & GPMVC                 & 51.44  & 46.89  & 43.91  & 28.23   & 20.04  & 15.48 & 58.39  & 52.76   & 45.29   \\
			\multicolumn{1}{c|}{}                           & UEAF                  & 78.22  & 77.24  & 69.31  & 70.71   & 68.25  & 55.13 & 87.41  & 87.07   & 77.07   \\
			\multicolumn{1}{c|}{}                           & IMVTSC-MVI            & 75.86  & 76.72  & 71.00  & 71.85   & 73.51  & 53.30 & 86.20  & 87.07   & 75.34   \\
			\multicolumn{1}{c|}{}                           & Ours                  & \textbf{80.17}  & \textbf{79.31}  & \textbf{73.28}  & \textbf{72.67}   & \textbf{73.64}  & \textbf{60.13} & \textbf{88.79 } & \textbf{89.66}   & \textbf{79.31}   \\ \hline
			\multicolumn{1}{c|}{\multirow{10}{*}{3Sources}} & BSV                   & 56.90  & 47.38  & 39.24  & 50.07   & 34.46  & 22.34 & 68.14  & 57.63   & 48.99   \\
			\multicolumn{1}{c|}{}                           & Concat                & 53.54  & 46.79  & 37.68  & 51.98   & 37.87  & 18.32 & 69.78  & 58.51   & 46.48   \\
			\multicolumn{1}{c|}{}                           & DAIMC                 & 56.33  & 52.43  & 50.73  & 52.98   & 49.07  & 41.64 & 68.99  & 67.21   & 63.56   \\
			\multicolumn{1}{c|}{}                           & IMSC-AGL              & 56.80  & 61.53  & 44.37  & 37.84   & 42.28  & 33.96 & 61.53  & 65.68   & 57.98   \\
			\multicolumn{1}{c|}{}                           & OPIMC                 & 59.76  & 62.82  & 53.90  & 59.24   & 57.15  & 48.87 & 75.14  & 76.92   & 67.73   \\
			\multicolumn{1}{c|}{}                           & PIC                   & 64.49  & 58.25  & 53.84  & 62.99   & 56.06  & 50.12 & 77.47  & 72.78   & 59.96   \\
			\multicolumn{1}{c|}{}                           & GPMVC                 & 48.24  & 44.50  & 42.01  & 34.82   & 30.44  & 28.15 & 60.47  & 58.58   & 57.4    \\
			\multicolumn{1}{c|}{}                           & UEAF                  & 62.6   & 55.62  & 52.78  & 56.47   & 52.06  & 45.19 & 75.50  & 71.95   & 67.69   \\
			\multicolumn{1}{c|}{}                           & IMVTSC-MVI            & 60.94  & 35.50  & 28.99  & 58.24   & 25.29  & 8.58  & 76.33  & 58.57   & 43.19   \\
			\multicolumn{1}{c|}{}                           & Ours                  & \textbf{66.27}  & \textbf{64.49}  & \textbf{56.03}  & \textbf{63.77}   & \textbf{60.82}  & \textbf{51.62} & \textbf{78.10}  & \textbf{77.51}   & \textbf{72.66}   \\ \hline
			\multicolumn{1}{c|}{\multirow{10}{*}{ORL}}      & BSV                   & 46.75  & 34.75  & 25.50  & 34.75   & 48.14  & 34.18 & 51.25  & 39.25   & 29.75   \\
			\multicolumn{1}{c|}{}                           & Concat                & 47.00  & 37.25  & 29.00  & 37.25   & 52.87  & 42.77 & 52.75  & 41.00   & 33.00   \\
			\multicolumn{1}{c|}{}                           & DAIMC                 & 52.50  & 49.25  & 31.75  & 49.25   & 66.56  & 44.49 & 59.00  & 54.25   & 29.50   \\
			\multicolumn{1}{c|}{}                           & IMSC-AGL              & 30.00  & 4.75   & 4.25   & 4.75    & 2.27   & 1.90  & 30.10  & 4.75    & 4.25    \\
			\multicolumn{1}{c|}{}                           & OPIMC                 & 55.50  & 54.75  & 32.50  & 54.75   & 72.34  & 45.36 & 60.50  & 60.75   & 30.00   \\
			\multicolumn{1}{c|}{}                           & PIC                   & 59.25  & 47.00  & 30.44  & 47.00   & 67.48  & 43.69 & 62.25  & 51.50   & 33.66 \\
			\multicolumn{1}{c|}{}                           & GPMVC                 & 53.23  & 51.36  & 31.69  & 51.369  & 68.13  & 43.72 & 61.53  & 58.73  & 32.10 \\
			\multicolumn{1}{c|}{}                           & UEAF                  & 41.77  & 20.2   & 15.37  & 20.20   & 28.61  & 21.92 & 45.55  & 23.50   & 17.80   \\
			\multicolumn{1}{c|}{}                           & IMVTSC-MVI            & 43.50  & 32.5   & 27.75  & 22.50   & 41.91  & 39.64 & 45.50  & 23.75   & 28.25   \\
			\multicolumn{1}{c|}{}                           & Ours                  & \textbf{60.00}  & \textbf{57.75}  & \textbf{34.5}   & \textbf{57.75}   & \textbf{73.69}  & \textbf{46.92} & \textbf{64.75}  & \textbf{61.25}   & \textbf{37.50}   \\ \hline
	\end{tabular}}
\end{table*}

\section{Experiments and Analysis of The Proposed MIMB}
In this section, various experiments are conducted with the proposed MIMB on diverse benchmark multi-view datasets with different incomplete settings. Moreover, comparative experiments are conducted with nine incomplete multi-view clustering methods to verify the superiority of MIMB. This section also provides a sensitivity analysis of diverse parameters, and a convergence analysis of the proposed MIMB. 
\subsection{Experimental settings}
We have provided the experimental settings in several parts, which are the dataset introduction and the details of the constructed incomplete multi-view data, comparison algorithms and evaluation metrics.
\subsubsection{Dataset Introduction}
Six common multi-view datasets are utilized in the experiments, namely, Caltech101-20, ORL, BDGP, 3Sources, BBCSport and Caltech101-7. The details are as follows:

\textbf{Caltech101-20} is the subset from the well-known multi-view dataset Caltech101 \cite{fei2004learning}, which contains 101 objects with 40-800 images in each object. Caltech101-20 includes 2386 samples with 20 classes that are extracted as six views, i.e., Gabor features, CENTRIST features, HOG features, WM features, GIST features and LBP features with 48, 254, 1984, 40, 512 and 928 dimensions, respectively.

\textbf{ORL} contains 400 facial images that were captured from 40 people of different ages and genders. In the experiments, this paper adopts four kinds of features as four views, which are extracted from a deep neural network with 400 dimensions.

\textbf{BDGP}, which includes 2500 Drosophila embryo samples with 5 categories from biological experiments, was extracted by medical instruments for gene expression research. In our experiments, the multi-view BDGP is divided into three views due to the different kinds of bag-of-words features. 

\textbf{3Sources} contains 169 news articles and six topics (i.e., health, business, sport, entertainment, politics and technology). The 169 samples are collected from three online sources, BBC representations with 3560 dimensions, Reuters representations with 3631 dimensions and The Guardian representations with 3068 dimensions.

\textbf{BBCSport} is collected from 737 news articles from common websites and is divided into five different classes (i.e., tennis, football, athletics, cricket and American football). Following the experiment in \cite{wen2022survey}, we select a subset that includes 116 samples with four different feature dimensions, 1991, 2063, 2113 and 2158.

\textbf{Caltech101-7} is a subset similar to the above Caltech101-20, which contains the same six feature dimensions. In contrast, Caltech101-7 includes 7 classes of objects and 1474 samples for each view.
\begin{table*}[tbp!]
	\centering
	\caption{Mean values of the performances with various methods on the  BDGP, Caltech-7 and Caltech-20 datasets with different incomplete ratios.} 
	\label{Table3}
	\scalebox{1}{
		\begin{tabular}{cc|ccc|ccc|ccc}
			\hline
			\multicolumn{2}{c|}{}                                                     & \multicolumn{3}{c|}{ACC} & \multicolumn{3}{c|}{NMI} & \multicolumn{3}{c}{Purity} \\ \hline
			\multicolumn{1}{c|}{Datasets}                     & Method/\ Missing rate & 0.1    & 0.3    & 0.5    & 0.1    & 0.3    & 0.5    & 0.1     & 0.3     & 0.5    \\ \hline
			\multicolumn{1}{c|}{\multirow{10}{*}{BGDP}}       & BSV                   & 27.60  & 28.76  & 27.90  & 3.09   & 3.55   & 3.42   & 27.68   & 28.80   & 28.08  \\
			\multicolumn{1}{c|}{}                             & Concat                & 26.16  & 40.80  & 33.80  & 2.17   & 13.04  & 10.19  & 26.2    & 40.80   & 33.84  \\
			\multicolumn{1}{c|}{}                             & DAIMC                 & 29.44  & 25.32  & 40.04  & 4.23   & 2.21   & 16.98  & 29.88   & 25.96   & 40.04  \\
			\multicolumn{1}{c|}{}                             & IMSC-AGL              & 29.12  & 33.84  & 35.20  & 4.39   & 7.64   & 7.01   & 30.12   & 35.20   & 35.76  \\
			\multicolumn{1}{c|}{}                             & OPIMC                 & 33.2   & 37.04  & 36.2   & 9.76   & 11.36  & 14.91  & 33.68   & 38.52   & 38.40  \\
			\multicolumn{1}{c|}{}                             & PIC                   & 32.92  & 22.2   & 23.52  & 10.17  & 1.97   & 1.94   & 33.16   & 22.78   & 23.95  \\
			\multicolumn{1}{c|}{}                             & GPMVC                 & 40.12  & 42.36  & 50.14  & 15.44  & 18.15  & 25.10  & 39.42   & 42.78   & 50.03  \\
			\multicolumn{1}{c|}{}                             & UEAF                  & 37.76  & 32.12  & 37.76  & 14.48  & 9.55   & 14.48  & 38.28   & 32.40   & 38.28  \\
			\multicolumn{1}{c|}{}                             & IMVTSC-MVI            & 32.72  & 35.32  & 46.60  & 9.86   & 12.33  & 22.29  & 33.08   & 35.68   & 46.60  \\
			\multicolumn{1}{c|}{}                             & Ours                  & \textbf{44.88}  & \textbf{45.52}  & \textbf{52.40}  & \textbf{21.43}  & \textbf{20.05}  & \textbf{26.55} & \textbf{46.36}   & \textbf{46.40}   & \textbf{53.84}  \\ \hline
			\multicolumn{1}{c|}{\multirow{10}{*}{Caltech-7}}  & BSV                   & 54.47  & 43.62  & 59.70  & 29.78  & 21.04  & 40.99  & 72.46   & 69.81   & 76.32  \\
			\multicolumn{1}{c|}{}                             & Concat                & 43.89  & 44.77  & 51.83  & 31.53  & 35.02  & 37.44  & 72.73   & 76.26   & 76.53  \\
			\multicolumn{1}{c|}{}                             & DAIMC                 & 41.79  & 47.55  & 42.33  & 37.62  & 37.79  & 42.08  & 83.38   & 81.34   & 84.87  \\
			\multicolumn{1}{c|}{}                             & IMSC-AGL              & 54.13  & 53.93  & 54.07  & 40.09  & 42.06  & 40.15  & 54.14   & 65.14   & 74.14  \\
			\multicolumn{1}{c|}{}                             & OPIMC                 & 51.28  & 53.12  & 55.42  & 41.71  & 45.51  & 44.59  & 83.11   & 83.62   & 84.19  \\
			\multicolumn{1}{c|}{}                             & PIC                   & 54.31  & 55.65  & 63.63  & 47.38  & 41.99  & 46.73  & 76.77   & 82.75   & 83.72  \\
			\multicolumn{1}{c|}{}                             & GPMVC                 & 53.65  & 54.79  & 60.79  & 37.69  & \textbf{45.66}  & 51.45  & 80.06   & 82.19   & 81.70  \\
			\multicolumn{1}{c|}{}                             & UEAF                  & 37.78  & 31.68  & 32.90  & 17.01  & 12.37  & 13.55  & 71.23   & 60.45   & 64.11  \\
			\multicolumn{1}{c|}{}                             & IMVTSC-MVI            & 52.71  & 55.54  & 64.85  & 44.59  & 49.83  & 55.13  & 86.50   & 82.86   & \textbf{88.74}  \\
			\multicolumn{1}{c|}{}                             & Ours                  & \textbf{54.92}  & \textbf{56.17}  & \textbf{65.92}  & \textbf{47.50}  & 43.13  & \textbf{54.78}  & \textbf{85.10}   & \textbf{83.72}   & 87.51  \\ \hline
			\multicolumn{1}{c|}{\multirow{10}{*}{Caltech-20}} & BSV                   & 35.71  & 36.42  & 39.23  & 34.09  & 37.46  & 43.76  & 56.41   & 57.96   & 61.53  \\
			\multicolumn{1}{c|}{}                             & Concat                & 29.97  & 30.60  & 37.30  & 38.18  & 40.33  & 45.00  & 62.99   & 63.50   & 68.69  \\
			\multicolumn{1}{c|}{}                             & DAIMC                 & 43.63  & 43.80  & 36.63  & 48.83  & 41.08  & 48.06  & 70.00   & 62.93   & 72.00  \\
			\multicolumn{1}{c|}{}                             & IMSC-AGL              & 33.45  & 33.53  & 33.32  & 10.06  & 15.17  & 20.06  & 33.45   & 33.53   & 33.45  \\
			\multicolumn{1}{c|}{}                             & OPIMC                 & 39.98  & 42.92  & 43.46  & 46.97  &\textbf{ 47.96}  & 51.14  & 70.06   & 61.67   & 71.85  \\
			\multicolumn{1}{c|}{}                             & PIC                   & 38.46  & 41.59  & 46.52  & 32.45  & 40.33  & 51.85  & 51.65   & 57.07   & 72.18  \\
			\multicolumn{1}{c|}{}                             & GPMVC                 & 30.66  & 33.89  & 36.69  & 35.58  & 38.25  & 42.17  & 55.43   & 61.21   & 68.57  \\
			\multicolumn{1}{c|}{}                             & UEAF                  & 23.89  & 24.10  & 27.75  & 21.79  & 22.86  & 26.59  & 48.62   & 49.16   & 50.80  \\
			\multicolumn{1}{c|}{}                             & IMVTSC-MVI            & 42.35  & 43.65  & \textbf{50.04}  & \textbf{48.29}  & 40.95  & \textbf{54.02}  & 57.26   & 60.37   & \textbf{75.10}  \\
			\multicolumn{1}{c|}{}                             & Ours                  & \textbf{44.17}  & \textbf{45.89}  & 49.26  & 47.41  & 43.18  & 52.14  & \textbf{70.16}   & \textbf{64.25}   & 72.30  \\ \hline
	\end{tabular}}
\end{table*}

In our experiments, to simulate real-world scenarios, two reverse strategies for constructing incomplete multi-view data are introduced with different incomplete settings. 

\textbf{(i) Incomplete case where each instance is randomly missing: }In our experiments, the BBCSport, 3Sources and ORL datasets are used to construct the random missing data case for each instance. We randomly removed $10\%$, $30\%$, and $50\%$ samples from every view, and preserved at least one view.

\textbf{(ii) Incomplete case in which random missing data with paired preserves: }Following the experimental settings in \cite{wang2022highly}, we randomly preserve $n_p$ paired instances, and then we generate a random index matrix to process the remaining instances. With the paired ratio ${{n_p}/n} =\{10\%, 30\%, 50\% \} $, we construct incomplete multi-view datasets for Caltech101-20, BDGP and Caltech101-7. For fairness, we repeated five different situations of the constructed data with random missing data and reported the average values as the final evaluation.

\subsubsection{Comparison Methods}
To assess the performance of the proposed MIMB, this paper chose several methods that are capable of completing the incomplete multi-view clustering task for comparison.
\begin{enumerate}[]
	\item BSV \cite{zhao2016incomplete}: BSV aims to uncover the underlying latent structure from multi-view data despite the presence of incomplete data, and then utilizes the k-means algorithm on the explored structure, ultimately yielding the optimal clustering results and reporting. In addition, BSV adopts the mean values of each view to impute the missing instances.
	\item Concat \cite{zhao2016incomplete}: Concat primarily merges the feature representations of multiple views into a unified view by concatenating them into a single vector and obtained the clustering results by adopting the k-means algorithm to the concatenated view. Similarly to BSV, Concat adopts a similar strategy to fill in missing views.
	\item DAIMC \cite{hu2018doubly}: DAIMC implements matrix factorization and sparse regression techniques to explore the consensus representation among multiple views, which fully utilizes existing instances via a doubly aligned strategy.
	\item IMSC-AGL \cite{wen2018incomplete}: IMSC-AGL constructs the graph of each view by subspace learning and imposes spectral constraints to learn the representation with low-dimensional features. For IMSC-AGL, the index matrix is introduced to cross the missing views.
	\item OPIMC \cite{hu2019one}: OPIMC provides an online strategy for large-scale datasets with chunks via the chunk training method. Moreover, OPIMC can directly obtain the cluster results by introducing two global statistics.
	\item PIC \cite{wang2019spectral}: The PIC first considers spectral perturbation in incomplete multi-view clustering tasks and provides a solid fusion criteria for multiple views. For PIC, the average similarity values are filled into the incomplete data for spectral clustering.
	\item GPMVC \cite{rai2016partial}: GPMVC proposed a partial-view algorithm for multiple incomplete views that exploits intrinsic geometry information from the data distribution of multiple views. For GPMVC, zero values are adopted to impute missing instances, and a matrix factorization strategy is utilized.
	
	\item UEAF \cite{wen2019unified}: The UEAF introduces a reconstruction term with a locality-preserved method to reconstruct incomplete data. This approach aimed to align multiple views and capture essential information from the complete data. 
	
	\item IMVTSC-MVI \cite{wen2021unified}: IMVTSC-MVI is a novel unified tensor framework that incorporates missing-view inference and high-order tensor information exploration and proposes the effective utilization of diverse features among multiple views for clustering.
\end{enumerate}

In our experiment, we utilize three well-known evaluation metrics, namely, accuracy (ACC), normalized mutual information (NMI) and purity, to assess the performance of the clustering task. These metrics provide a comprehensive evaluation of the clustering quality. For all the methods that were compared, the higher the value of the evaluation metric is, the better the clustering performance of the algorithm.

\begin{figure*}[tbp!]
	\centering
	
	\setlength{\belowcaptionskip}{-1mm}
	\vspace{-0.35cm}
	\subfigtopskip=-1pt 
	\subfigbottomskip=-1pt 
	\subfigcapskip=-5pt 
	\subfigure[BBCSport]{
		\includegraphics[width=4.3cm]{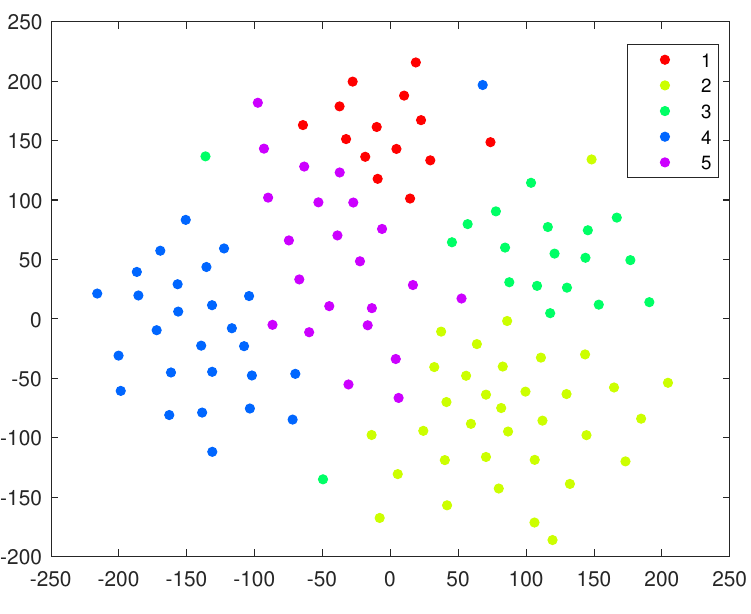}}
	\subfigure[3Sources]{
		\includegraphics[width=4.3cm]{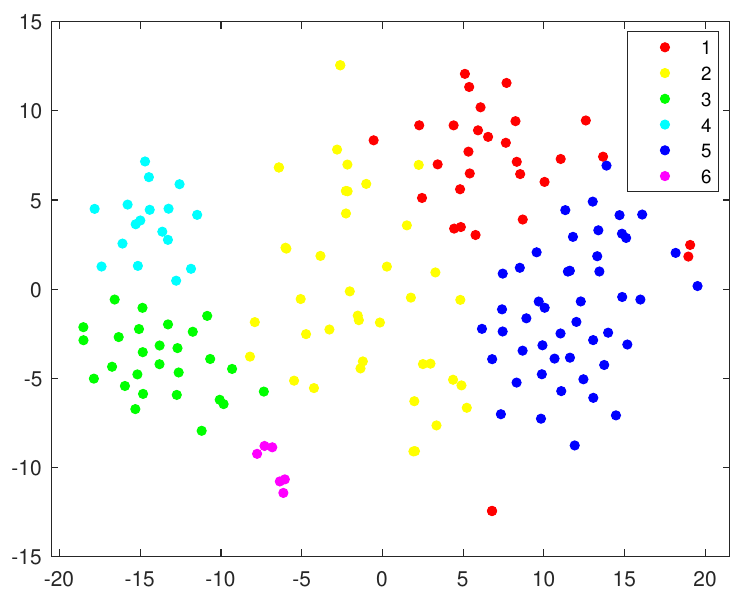}}
	\subfigure[BDGP]{
		\includegraphics[width=4.3cm]{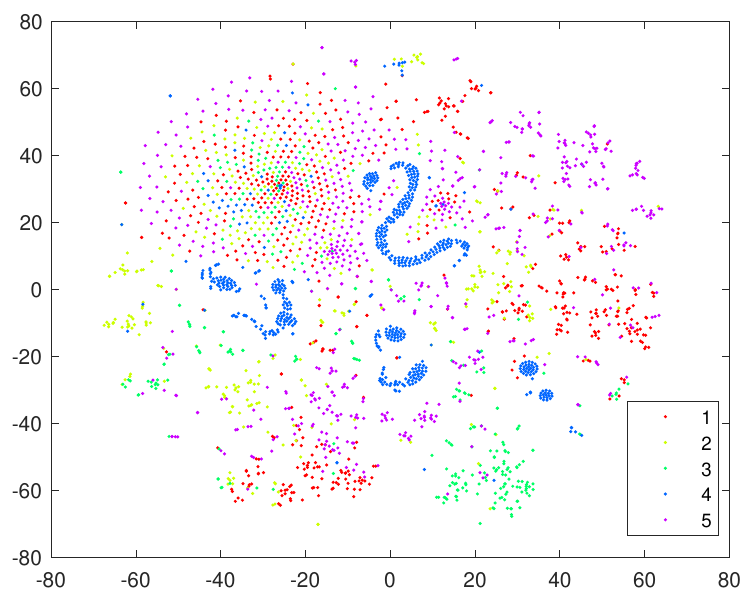}}
	\subfigure[Caltech-7]{
		\includegraphics[width=4.3cm]{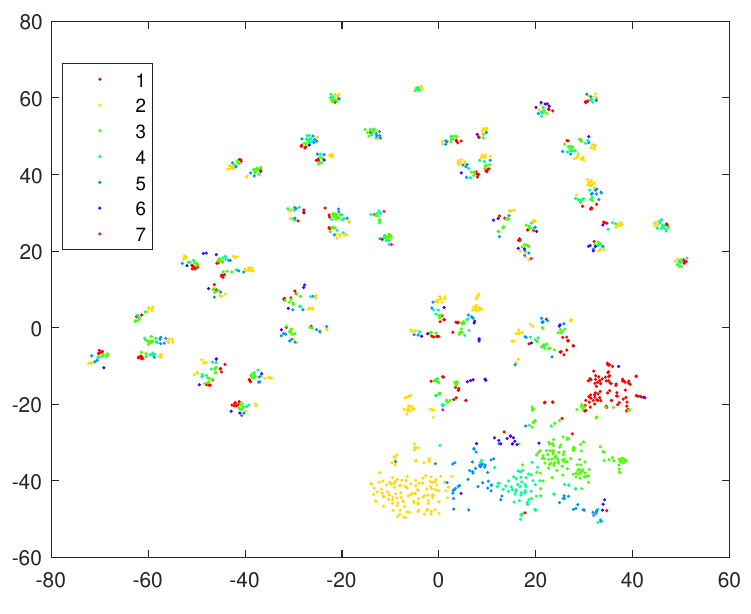}}			
	\caption{ T-SNE for visualizing the feature space of the final clustering representation on the BBCSport, 3Sources, BDGP and Caltech-7 datasets with a 30\% incomplete rate, respectively  }
	\label{tsne}
\end{figure*}

\subsection{Experimental Results and Analysis}
The summarized experimental results on the constructed incomplete datasets are shown in Table. \ref{Table2} and Table. \ref{Table3}. The best performances of the mean values of ACC (\%), NMI (\%) and purity (\%) are represented by the bold values. Above all, by comparing the experimental results on the constructed incomplete datasets, it is obvious that MIMB almost outperforms the other methods and achieves the best performance across various metrics. Specifically, on the well-known BBCSport dataset, with a 30\% missing data rate, the improvements in the accuracy of MIMB over the suboptimal method (i.e., UEAF) are 2\% in ACC, 5\% in NMI and 2\% in purity. Moreover, on the facial recognition dataset ORL with a 50\% missing rate, the performance of MIMB performs significantly better than the recent method (IMVTSC-MVI) in the above measures 7\% ACC, 7\% NMI, 9\% Purity. Meanwhile, it is not difficult to observe that as the missing rate increases, the values of ACC, NMI, and Purity in all methods continue to decrease. This further indicates that achieving satisfactory clustering results with incomplete multi-view data becomes more challenging as the number of missing instances increases.

To showcase the universality and effectiveness of the MIMB for real-world incomplete situations, we adopt a second strategy to construct incomplete datasets for experimenting with three missing ratios, and the results are listed in Table. \ref{Table3}. For the partial preserving datasets, our proposed method also exhibits competitive performances as compared to the different incomplete multi-view methods. Taking the experiments on the BDGP gene expression dataset, and the widely used Caltech-20 dataset, as examples, the proposed MIMB usually achieves satisfactory performances with different preservation rates. Specifically, in contrast with the 2nd ranked IMVTSC-MVI on Caltech-20 with a 30\% preserving ratio, MIMB has increased by 2\%, 3\% and 4\% in ACC, NMI and Purity, respectively. It is noteworthy that in 10\% preserving ratio on BDGP, the proposed method has significant improvements over IMVTSC-MVI method are 12\% in ACC, 12\% in NMI and 13\% in Purity. This fully testifies the clear effectiveness of MIMB method for recovering missing views and embedding the manifold structure for consensus representation. Besides, it is obvious that the evaluation matrices are increasing with the preserving ratios improving, which demonstrate more available instances will lead better clustering performance.

Compared with previous partially aligned methods (i.e., BSV, Concat, PIC, etc.) that have not recovered incomplete views, MIMB can obtain better clustering results in diverse situations by flexibly combining the existing information. This is because the proposed method is designed to explore latent representations among all the views, and then simultaneously learn the consensus representation and recover the missing instances. Furthermore, unlike the existing recovery-based methods (i.e., UEAF, IMVTSC-MVI, etc.), MIMB considers the unbalanced distribution between the recovered data and original data, and embeds the manifold structure into a consensus representation. Specifically, taking the clustering performances on BBCSport and 3Sources as examples, our proposed method obtains a satisfactory performance as compared to UEAF and IMVTSC-MVI, and the corresponding improvement rates are 2\%, 5\%, 4\% and 6\%, respectively, in terms of the ACC with a 0.1 missing rate. To be more intuitive, t-SNE figures were constructed to clearly visualize the clustering performance of MIMB. As illustrated in Fig. \ref{tsne}, we utilize t-SNE on BBCSport, 3Sources, BDGP and Caltech-7 with 30\% recovered datasets with view concatenation data, in which the different colors represent the different classes of data. Obviously, from the distribution of the recovered samples, our proposed method has achieved great results because of the manifold structure embedding and reverse representation regularization.

\begin{figure*}[tbp!]
	\centering
	
	\setlength{\belowcaptionskip}{-1mm}
	\vspace{-0.35cm} 
	\subfigtopskip=-1pt 
	\subfigbottomskip=-1pt 
	\subfigcapskip=-5pt 
	\subfigure[BBCSport]{
		\includegraphics[width=4.3cm]{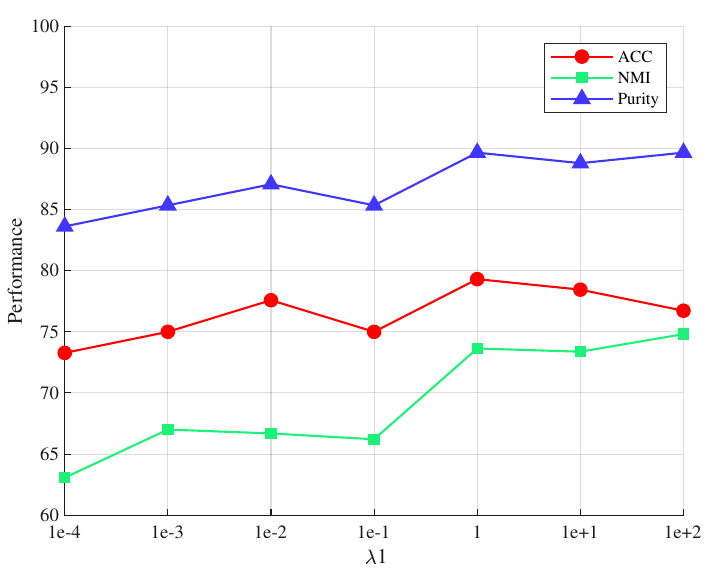}}
	\subfigure[3Sources]{
		\includegraphics[width=4.3cm]{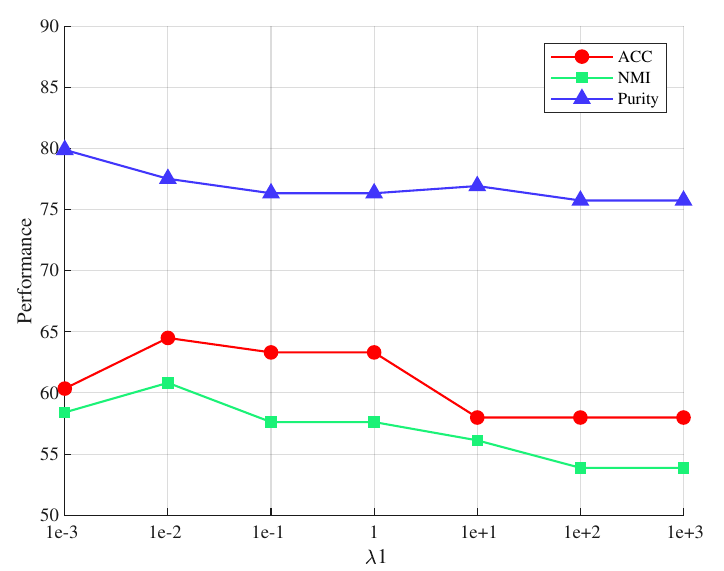}}
	\subfigure[BDGP]{
		\includegraphics[width=4.3cm]{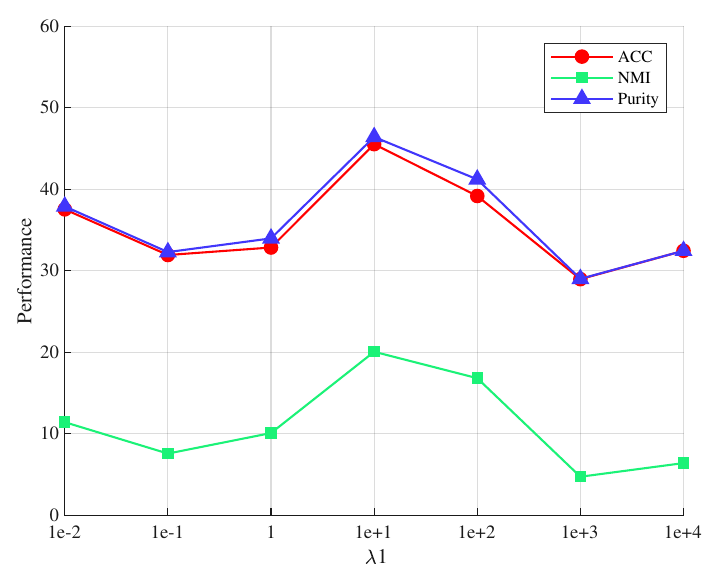}}
	\subfigure[Caltech-7]{
		\includegraphics[width=4.3cm]{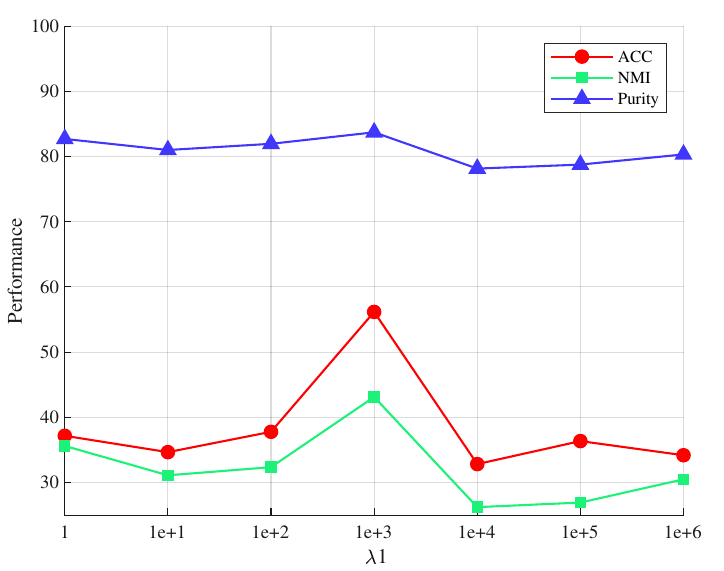}}			
	\caption{$\lambda_1 $ parameter adjustments on the BBCSport, 3Sources, BDGP and Caltech-7 datasets}
	\label{lambda1}
\end{figure*}
\begin{figure*}[tbp!]
	\centering
	\setlength{\belowcaptionskip}{-1mm}
	\vspace{-0.35cm} 
	\subfigtopskip=-1pt 
	\subfigbottomskip=-1pt 
	\subfigcapskip=-5pt 
	\subfigure[BBCSport]{
		\includegraphics[width=4.3cm]{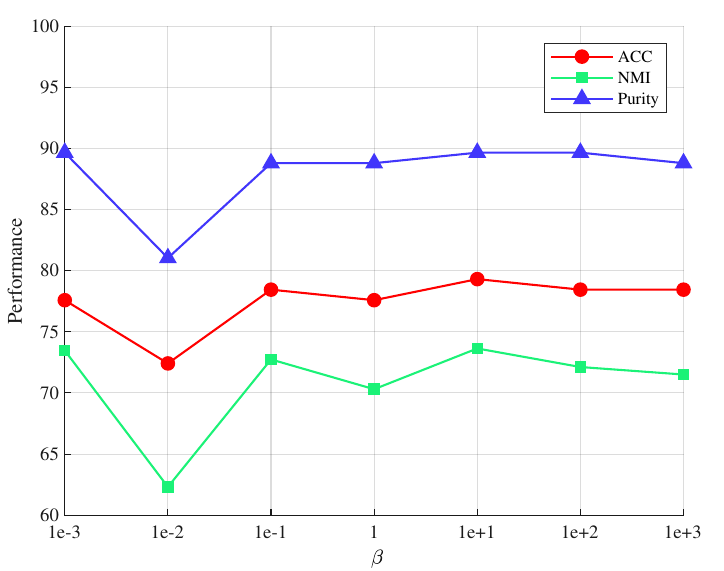}}
	\subfigure[3Sources]{
		\includegraphics[width=4.3cm]{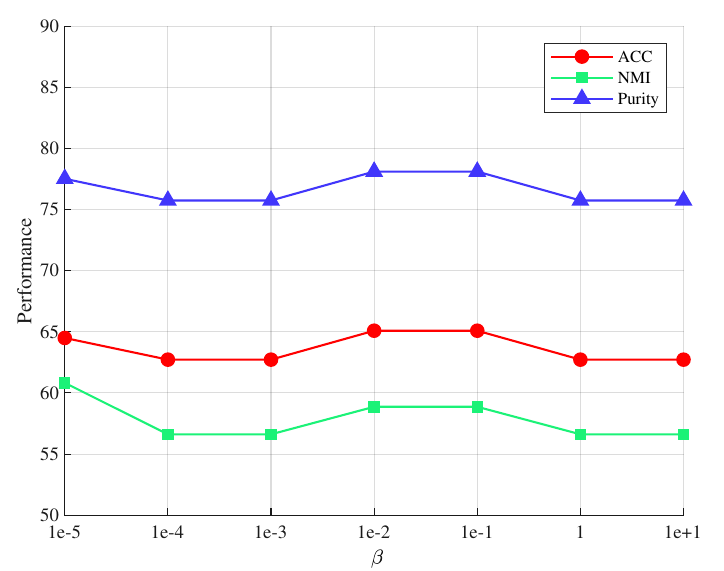}}
	\subfigure[BDGP]{
		\includegraphics[width=4.3cm]{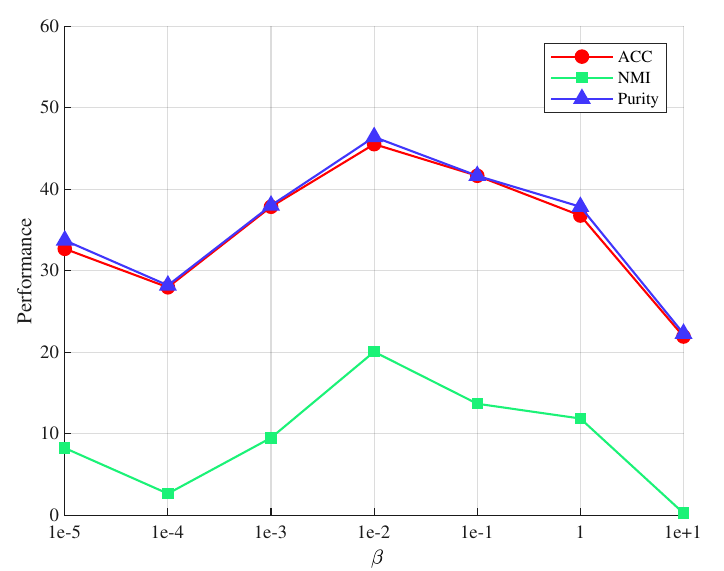}}
	\subfigure[Caltech-7]{
		\includegraphics[width=4.3cm]{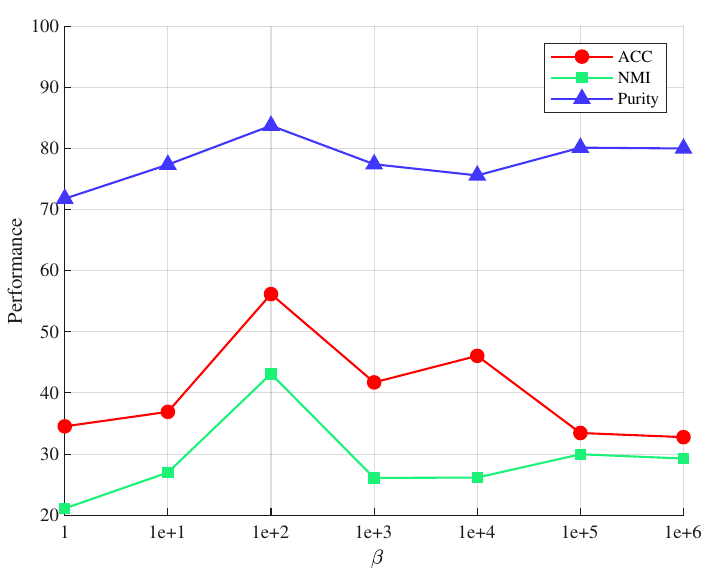}}			
	\caption{ $\beta$ parameter adjustments on the BBCSport, 3Sources, BDGP and Caltech-7 datasets}
	\label{beta}
\end{figure*}
\begin{figure*}[tbp!]
	\centering
	\setlength{\belowcaptionskip}{-1mm}
	\vspace{-0.35cm} 
	\subfigtopskip=-1pt
	\subfigbottomskip=-1pt 
	\subfigcapskip=-5pt 
	\subfigure[BBCSport]{
		\includegraphics[width=4.3cm]{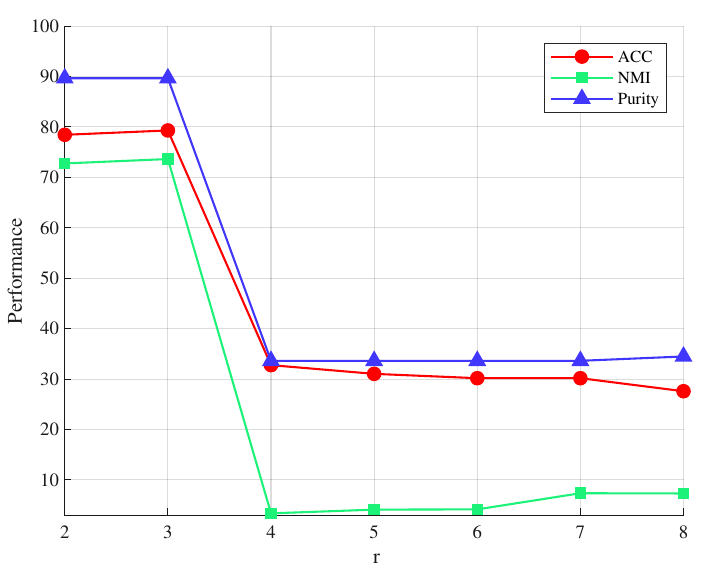}}
	\subfigure[3Sources]{
		\includegraphics[width=4.3cm]{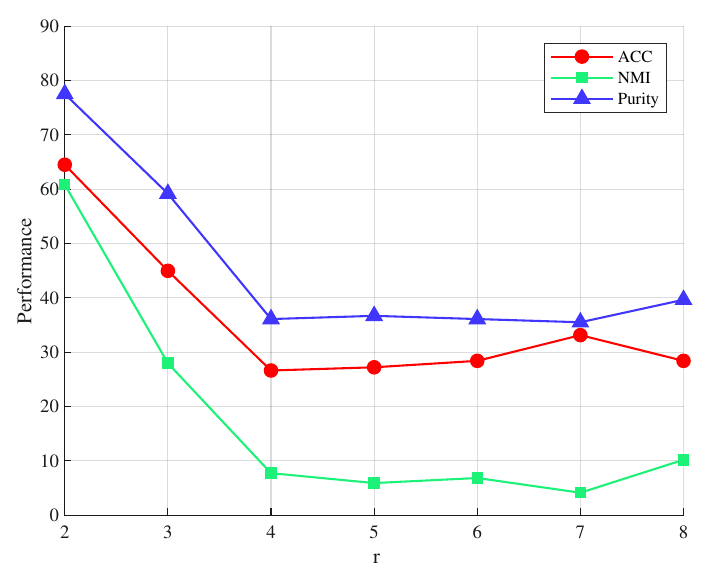}}
	\subfigure[BDGP]{
		\includegraphics[width=4.3cm]{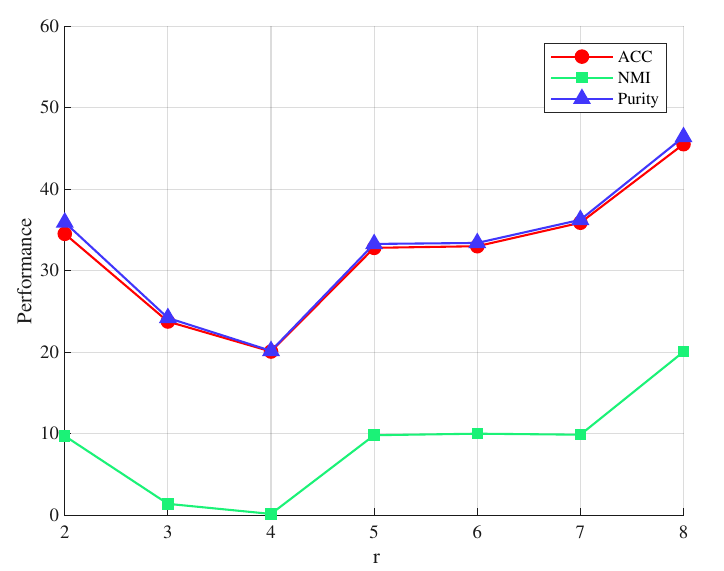}}
	\subfigure[Caltech-7]{
		\includegraphics[width=4.3cm]{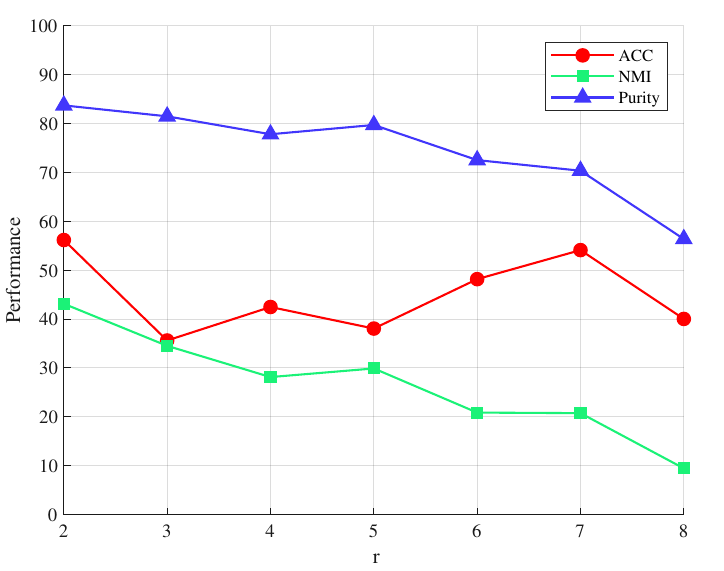}}			
	\caption{ $r$ parameter adjustments on the BBCSport, 3Sources, BDGP and Caltech-7 datasets, respectively}
	\label{r}
\end{figure*}

\subsection{ Parameter Sensitivity and Convergence Analysis }
This paper conducts experiments on four penalty parameters, $\lambda_1$, $\lambda_2$, $\lambda_3$ and $\beta$, and a smoothing parameter, $r$, in this section, which mainly focused on analyzing the parameter sensitivity of our proposed MIMB. First, we conducted diverse experiments with different selections of $\lambda_1$, $\beta$ and $r$ on the BBCSport, 3Sources, BDGP and Caltech-7 datasets. The constructed incomplete datasets are selected at a 30\% missing/preserving ratio for these experiments. The trends of $\lambda_1$, $\beta$ and $r$ were observed intuitively with the calculated performances of the cluster results, and are displayed using line charts. As shown in Fig. \ref{lambda1}, we tune $\lambda_1$ in the range of $ [1e-2, 1e+4]$, and the different change trends on the 4 incomplete datasets clearly show that the values of $\lambda_1$ affect the cluster performances. In addition, the trends of $\beta$ are provided in Fig. \ref{beta}. With the changing values of $\beta$, the parameter sensitivity of the MIMB can be analyzed by calculating the performance metrics ACC, NMI and purity. It is obvious that the proposed MIMB can achieve the ideal results with these datasets when $\beta $ is selected to a modest value. Then, as shown in Fig. \ref{r}, we conducted experiments with different selections of $r$ in the range of $[2,8]$. Notably, an optimal selection of $r$ can lead our proposed MIMB to obtain a satisfactory performance, which is the weight parameter used to effectively balance multiple views.

\begin{figure*}[tbp!]
	\centering
	\setlength{\belowcaptionskip}{-1mm}
	\vspace{-0.35cm} 
	\subfigtopskip=-1pt 
	\subfigbottomskip=-1pt 
	\subfigcapskip=-5pt 
	\subfigure[BBCSport]{
		\includegraphics[width=4.3cm]{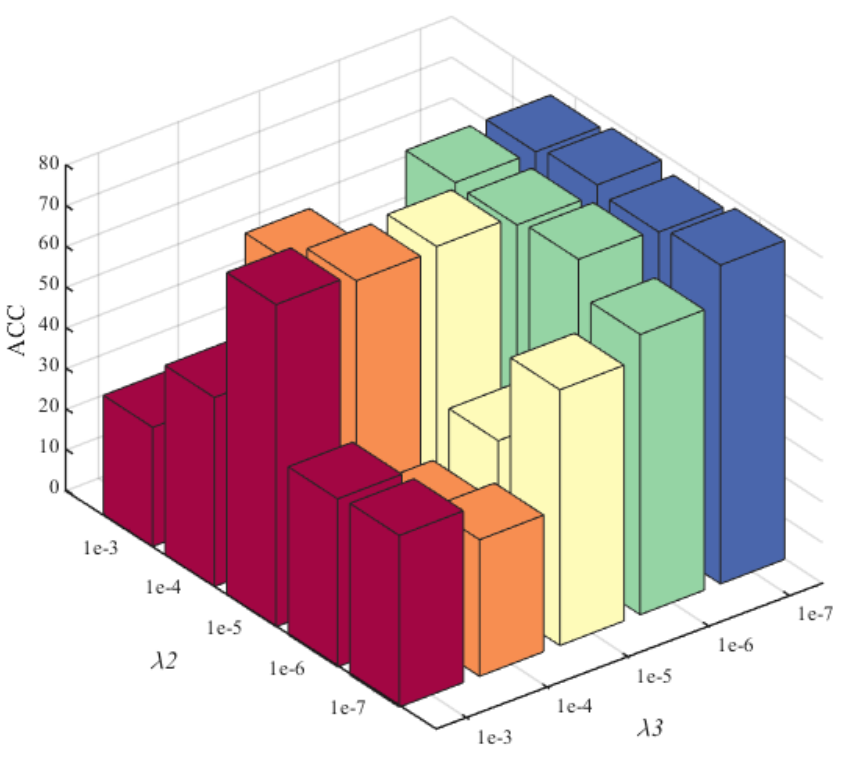}}
	\subfigure[3Sources]{
		\includegraphics[width=4.3cm]{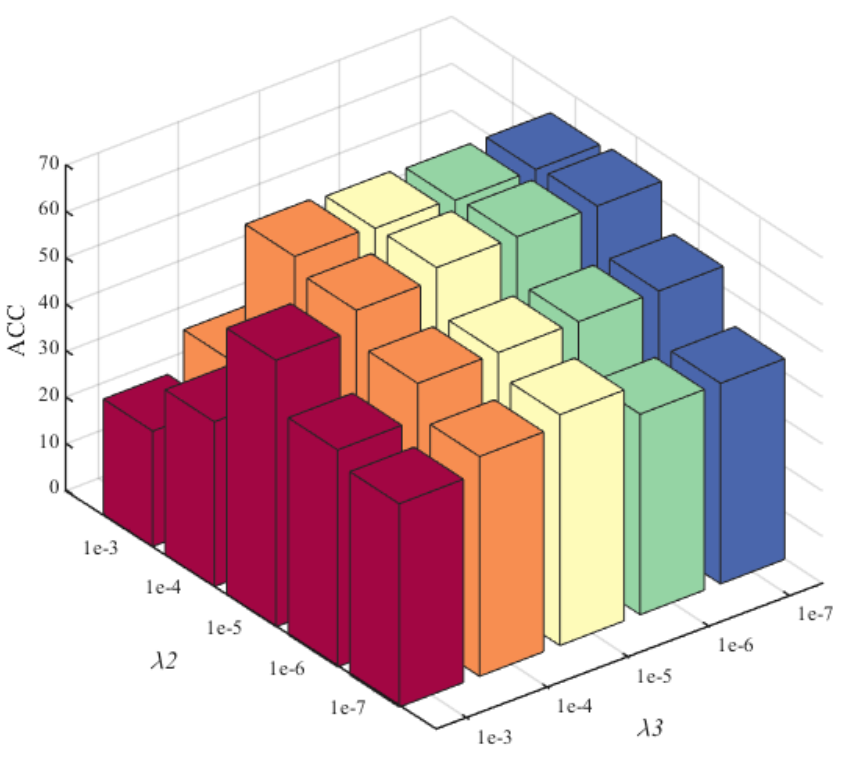}}
	\subfigure[BDGP]{
		\includegraphics[width=4.3cm]{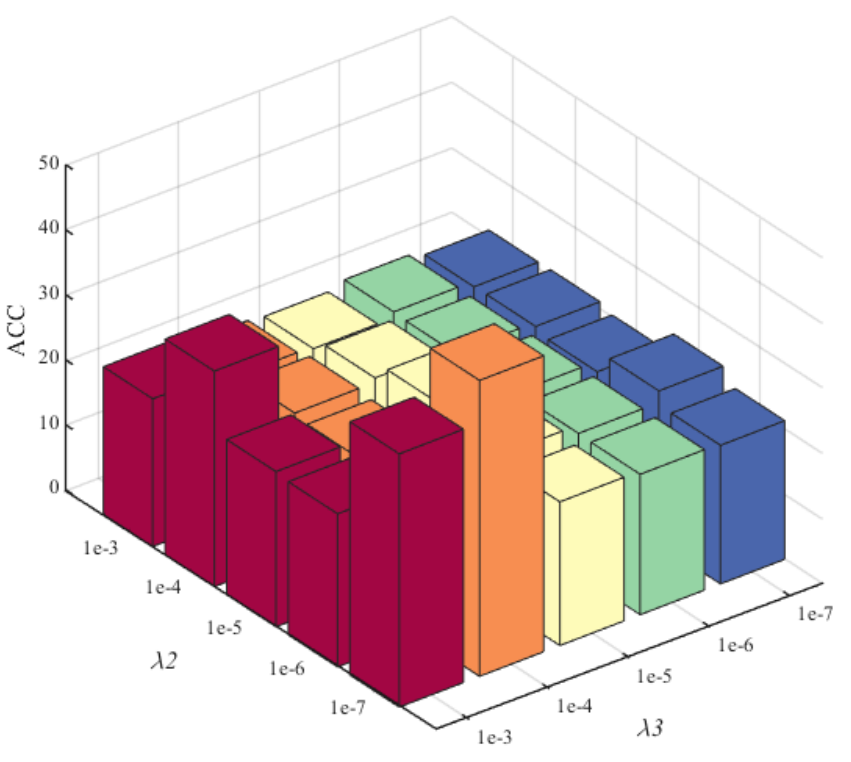}}
	\subfigure[Caltech-7]{
		\includegraphics[width=4.3cm]{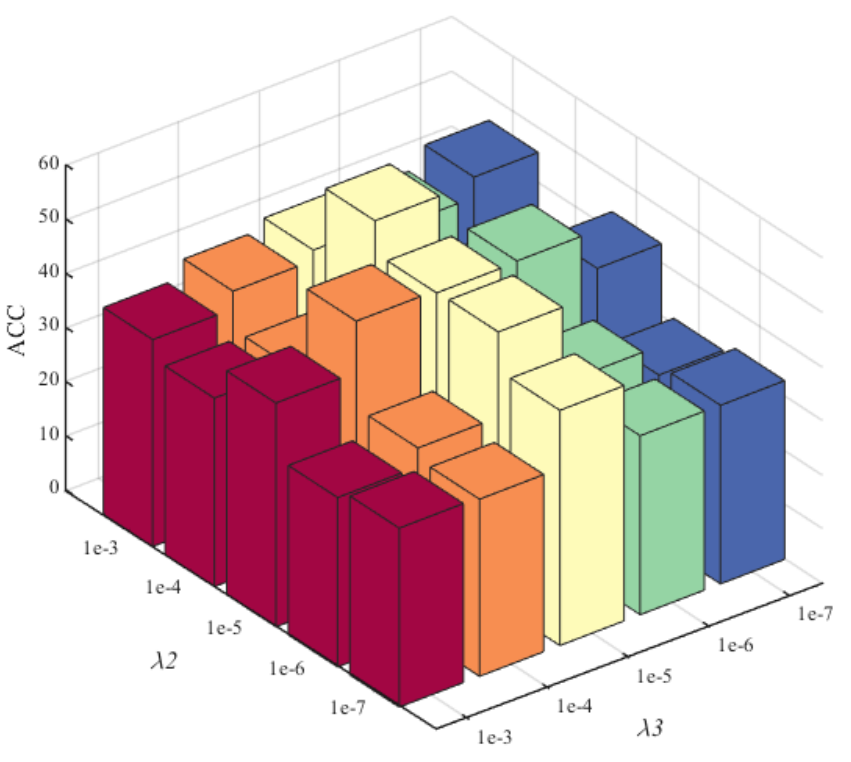}}			
	\caption{ Parameter adjustments for $\lambda_2$ and $\lambda_3$ on the BBCSport, 3Sources, BDGP and Caltech-7 datasets, respectively}
	\label{3D}
\end{figure*}

To demonstrate the effectiveness of manifold embedding, the parameter sensitivity experiments conducted with different settings of $\lambda_2$ and $\lambda_3$ on these incomplete datasets are illustrated in Fig. \ref{3D}. MIMB adopts $\lambda_2$ and $\lambda_3$ as the weights of the manifold embedding constraints to balance the loss values. Specifically, the clustering results obtained by adjusting the values of $\lambda_2$ and $\lambda_3$ are generated as three-dimensional statistical figures. The 3D statistical figures on the BBCSport, 3Sources, BDGP and Caltech-7 datasets with missing/preserving ratios of 30\% are shown in Fig. \ref{3D}. It is obvious that the best selection of $\lambda_2$ and $\lambda_3$ can achieve the best cluster performance for our proposed method. Specifically, we set $\lambda_2$ and $\lambda_3$ both in the range $[1e-3,1e-7]$, and the best choices of BBCSport are $\lambda_2=1e-5$ and $\lambda_3=1e-5$. 

\begin{figure}[tbp!]
	\centering
	
	\setlength{\belowcaptionskip}{-1mm}
	\vspace{-0.35cm} 
	\subfigtopskip=-1pt
	\subfigbottomskip=-1pt 
	\subfigcapskip=-5pt 
	\subfigure[BBCSport]{
		\includegraphics[width=4.2cm]{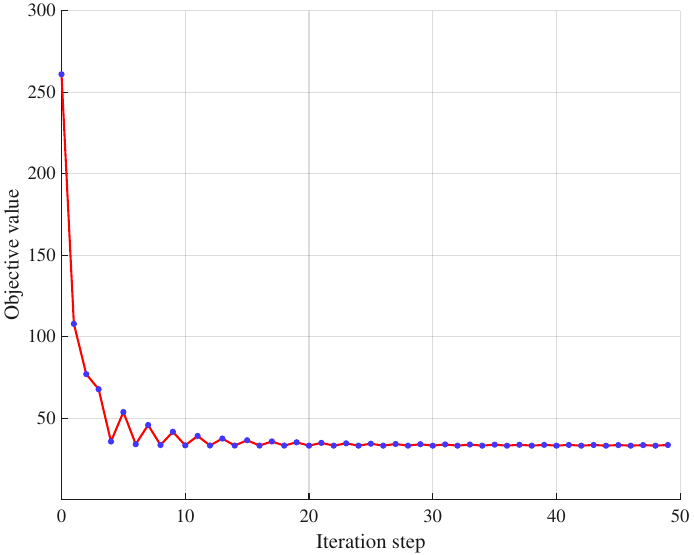}}
	\subfigure[3Sources]{
		\includegraphics[width=4.2cm]{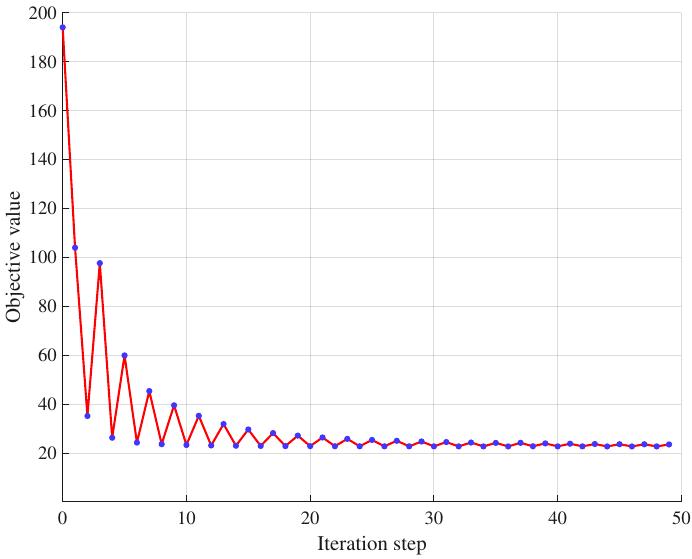}}
	\subfigure[BDGP]{
		\includegraphics[width=4.2cm]{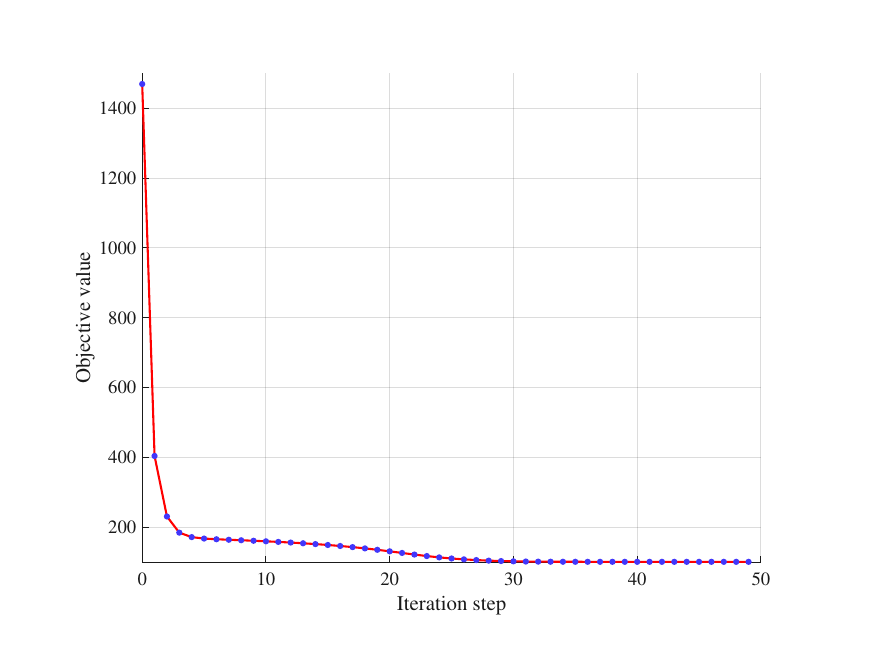}}
	\subfigure[Caltech-7]{
		\includegraphics[width=4.2cm]{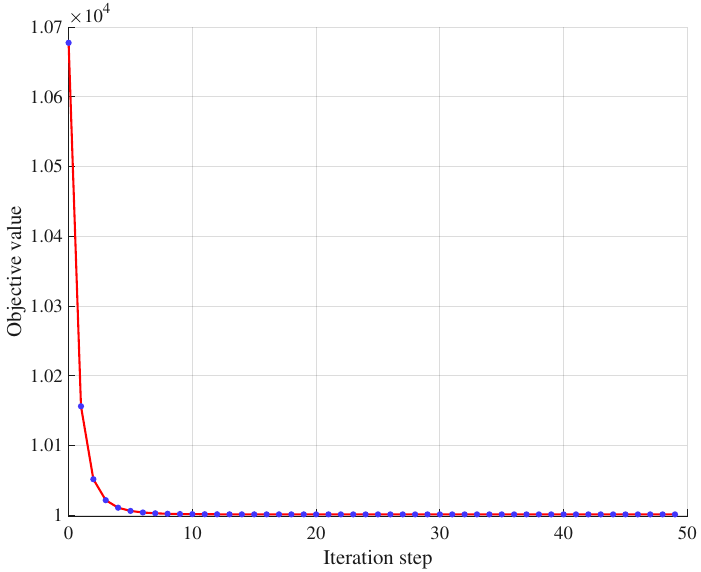}}			
	\caption{Convergence results of the MIMB on the BBCSport, 3Sources, BDGP and Caltech-7 datasets }
	\label{con}
\end{figure}

In this section, convergence analysis experiments were performed to verify the convergence of the proposed optimization algorithm, and the results are summarized in Fig. \ref{con}. Specifically, these figures depict the decreasing trend of the objective values as the number of iterations increases on the BBCSport, 3Sources, BDGP and Caltech-7 datasets. The curve of the objective value decreases monotonically until it reaches a stable value as the number of iteration steps increases. Furthermore, the effectiveness of our proposed MIMB for the above datasets is briefly demonstrated by the phenomenon that the function objective values converge quickly after 5-10 iteration steps.

\section{Conclusion}
In this paper, we propose a novel recovery-based IMVC method termed MIMB. Unlike the existing recovery-based methods, MIMB fully considers the mutual influence between the recovered data and existing data, which aims to reduce the gap distribution during the recovery process, flexibly explores the consistency information from incomplete data and then embeds the manifold structure into the consensus representation. Importantly, the addition of reverse representation regularization for recovery-based consensus representation learning aims to reduce noise and ensure the effective exploration of the latent consistency information from the recovered data. Moreover, MIMB effectively preserves the manifold structure among multiple views in the consensus representation, which is essential for obtaining the final clustering results. In addition, the proposed MIMB also utilizes an autoweighted strategy for balancing different views, which attempts to fully integrate the underlying information from various views. Compared with the recovery-based methods, MIMB can reasonably balance the complete data and recovered data for clustering. Through comparative experiments with 6 widely used multi-view datasets, MIMB has been proven to be superior to several state-of-the-art incomplete multi-view clustering methods. In the future, we will attempt to improve the suitability of the proposed MIMB for cross-modal retrieval tasks and reduce the time complexity and the  number of parameters.

\section*{Acknowledge}
This work was supported in part by the National Natural Science Foundation of China Grants 62172136, U21A20470, U1936217, 62002041 and 62176037; the Liaoning Fundamental Research Funds for Universities Grant LJKQZ2021010; the Liaoning Doctoral Research Startup Fund Project Grant 2021-BS-075; and the Dalian Science and Technology Innovation Fund 2022JJ12GX019, 2021JJ12GX028. 

\bibliographystyle{IEEEtran}

\bibliography{mybibfile}

\begin{IEEEbiography}[{\includegraphics[width=1in, height=1.35in]{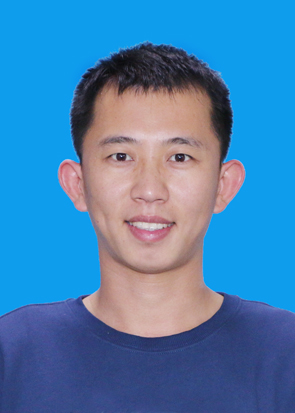}}]{Huibing Wang}
	received a Ph.D. degree in the School of Computer Science and Technology, Dalian University of Technology, Dalian, in 2018. During 2016 and 2017, he was a visiting scholar at the University of Adelaide, Adelaide, Australia. Now, he is an Associate Professor at Dalian Maritime University, Dalian, Liaoning, China.
	
	He has authored and coauthored more than 80 papers in some famous journals and conferences. Furthermore, he serves as reviewers for IEEE TPAMI, IEEE TKDE, IEEE TIP, ACM TOIS, IEEE TNNLS, IEEE TMM, IEEE TCYB, ACM TOMM, etc., and as SPC or PC for CVPR, ACM MM, AAAI, IJCAI, ECCV, ICME, et al.. His research interests include computer vision and machine learning.
\end{IEEEbiography}

\begin{IEEEbiography}[{\includegraphics[width=1in, height=1.35in]{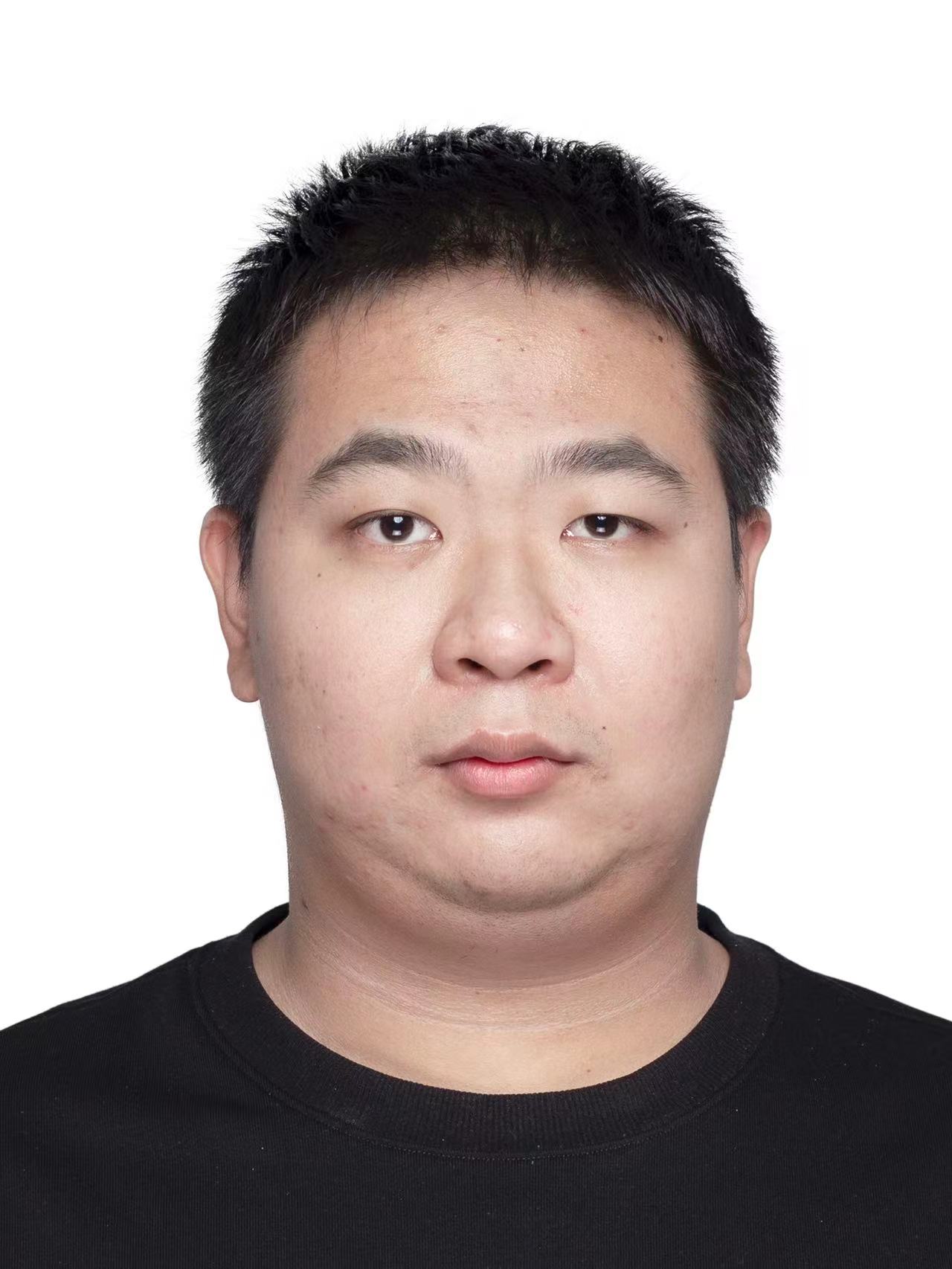}}]{Mingze Yao}
	received the B.S. degree from Jiangsu University. Now, he	is a Master candidate in Dalian Maritime University. His research interests include multi-view learning and person search.
\end{IEEEbiography}

\begin{IEEEbiography}[{\includegraphics[width=1in, height=1.35in]{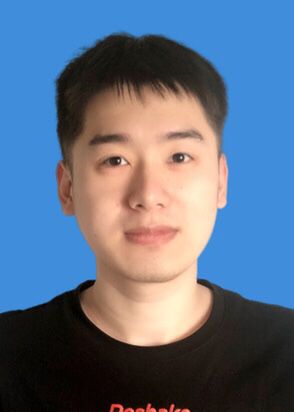}}]{Yawei Chen}
	received the B.S. degree from Hebei Agricultural University. Now, he is a Master candidate in Dalian Maritime University. His research interest is multi-view learning.
\end{IEEEbiography}

\begin{IEEEbiography}[{\includegraphics[width=1in, height=1.35in]{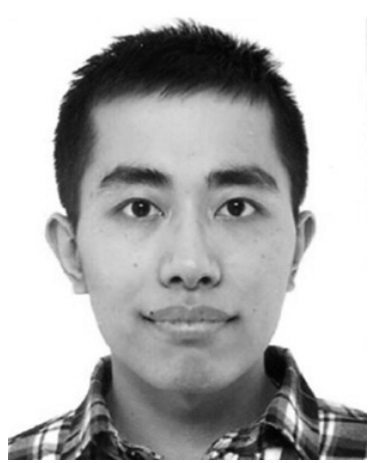}}]{Yunqiu Xu}
	received the B.S. degree in polymer materials from Zhejiang University, Hangzhou, China, in 2014, and the M.S. degree in information technology from the University of New South Wales, Sydney, NSW, Australia, in 2018. He is currently pursuing the Ph.D. degree in computer science with the University of Technology Sydney, Sydney. His research focuses on deep learning, reinforcement learning, natural language processing, and intelligent transportation systems
\end{IEEEbiography}

\begin{IEEEbiography}[{\includegraphics[width=1in, height=1.35in]{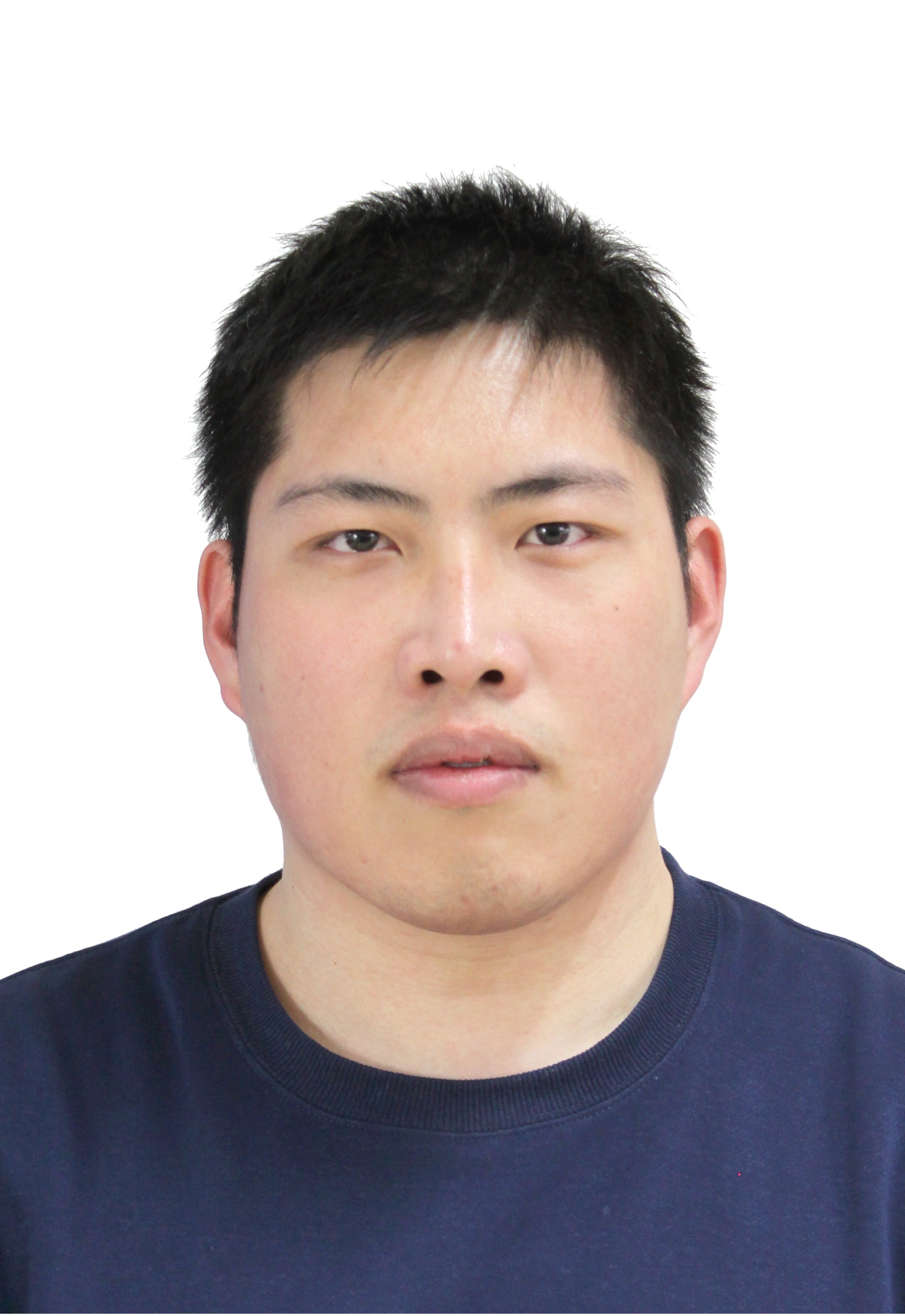}}]{Haipeng Liu}
 is currently a Ph.D. candidate at the Hefei University of Technology, Hefei, China. His research interests include computer vision, image inpainting. He has published several research papers including CVPR, ACM MM. 
\end{IEEEbiography}

\begin{IEEEbiography}[{\includegraphics[width=1in, height=1.35in]{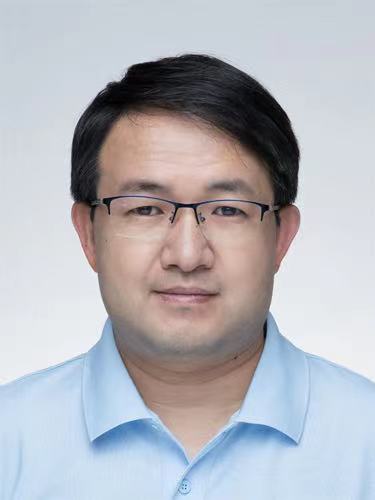}}]{Wei Jia}
received the B.Sc. degree in informatics from Central China Normal University, Wuhan, China, in 1998, the M.Sc. degree in computer science from the Hefei University of Technology, Hefei, China, in 2004, and the Ph.D. degree in pattern recognition and intelligence systems from the University of Science and Technology of China, Hefei, in 2008. He was a Research Assistant Professor and an Associate Professor with the Hefei Institutes of Physical Science, Chinese Academy of Science, from 2008 to 2016. He is currently a Professor with the School of Computer and Information, Hefei University of Technology. His research interests include computer vision, biometrics, pattern recognition, and image processing.
\end{IEEEbiography}

\begin{IEEEbiography}[{\includegraphics[width=1in, height=1.35in]{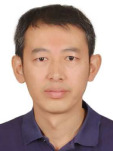}}]{Xianping Fu} is a Full Professor at Dalian Maritime University, China. He received a Ph.D. degree from Dalian Maritime University in 2005. He previously worked as a postdoc at Tsinghua University in 2008 and senior research fellow at Harvard University. His major research interests are image processing for content recognition, multimedia technology and underwater robot vision.
	
He has authored over 100 journal and conference papers in these areas, which have been published in IJCAI, TMM, TITS, TCSVT, TVT, ICME, ICMR, OCEANs, etc. He won the American RPB international scholar research award in 2009. His group was included in the Liaoning Revitalization Talents Program. Now, he is working as the dean of the College of Information Science and Technology in DMU and director of the Liaoning Underwater Robot Engineering Research Center.
\end{IEEEbiography}

\begin{IEEEbiography}[{\includegraphics[width=1in, height=1.35in]{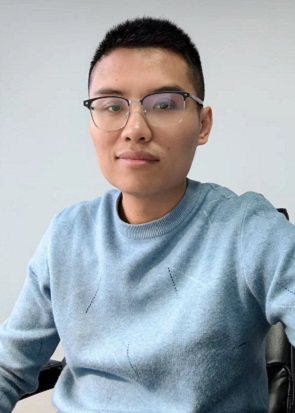}}]{Yang Wang} is currently a Hubing Professor at Hefei University of Technology, China. He has been selected by National Ten-thousands Talent Program of China. 
	
So far, He has published more than 100 research papers, such as IEEE Transactions on Pattern Analysis and Machine Intelligence, Artificial Intelligence (Elsevier), International Journal of Computer Vision, IEEE Trans. Image Processing, CVPR, ECCV, KDD, SIGIR, AAAI, IJCAI, ACM Multimedia, Machine Learning (Springer), ACM Trans. Information Systems, ACM Trans. Knowledge Discovery from Data, IEEE TKDE and VLDB Journal. Yang Wang is currently an Associate Editor for ACM Trans. Information Systems. His research has gained more than 6000 google scholar citations with H-index 38. 
\end{IEEEbiography}

\end{document}